\documentclass{cambrian}
\usepackage[utf8]{inputenc} %
\usepackage[T1]{fontenc}    %
\usepackage{hyperref}       %
\usepackage{url}            %
\usepackage{booktabs}       %
\usepackage{amsfonts}       %
\usepackage{nicefrac}       %
\usepackage{microtype}      %

\usepackage{mwe}
\usepackage{graphicx}
\usepackage{amssymb}
\usepackage{tabularray}
\usepackage{adjustbox}
\usepackage{colortbl}
\usepackage{array}
\usepackage{multirow}
\usepackage{makecell}
\usepackage{bbm}
\usepackage{collcell,xfp}
\usepackage{pgf}
\usepackage{tikz}
\usepackage[most]{tcolorbox}
\usepackage{csquotes}
\usepackage[noorphans,vskip=1em,leftmargin=1em]{quoting}
\usepackage{pifont} %
\usepackage{enumitem} %
\usepackage{tcolorbox} %
\usepackage{forest}
\usepackage{wrapfig}
\usepackage{caption}
\usepackage{subcaption}
\usepackage{longtable}
\usepackage{algpseudocode}
\usepackage{amsmath}
\usepackage[ruled,vlined,linesnumbered]{algorithm2e}
\usepackage[capitalize]{cleveref}
\crefname{algocf}{Algorithm}{Algorithms}
\Crefname{algocf}{Algorithm}{Algorithms}
\usepackage[percent]{overpic}
\usepackage{xspace}
\usepackage[normalem]{ulem}  %
\usepackage{tocloft}  %
\usepackage{natbib}  %
\usepackage{mathtools}
\usepackage{listings}

\usepackage{fontawesome}
\newcommand{\huggingface}{\raisebox{-1.5pt}{\includegraphics[height=1.05em]{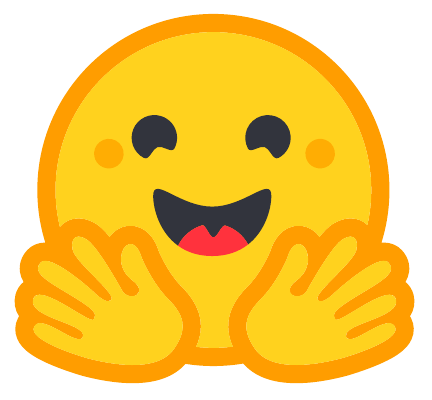}}\xspace}
\newcommand{\github}{\raisebox{-1.5pt}{\includegraphics[height=1.05em]{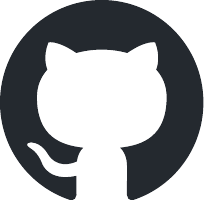}}\xspace}
\newcommand{\worldwideweb}{\raisebox{-1.5pt}{\includegraphics[height=1.05em]{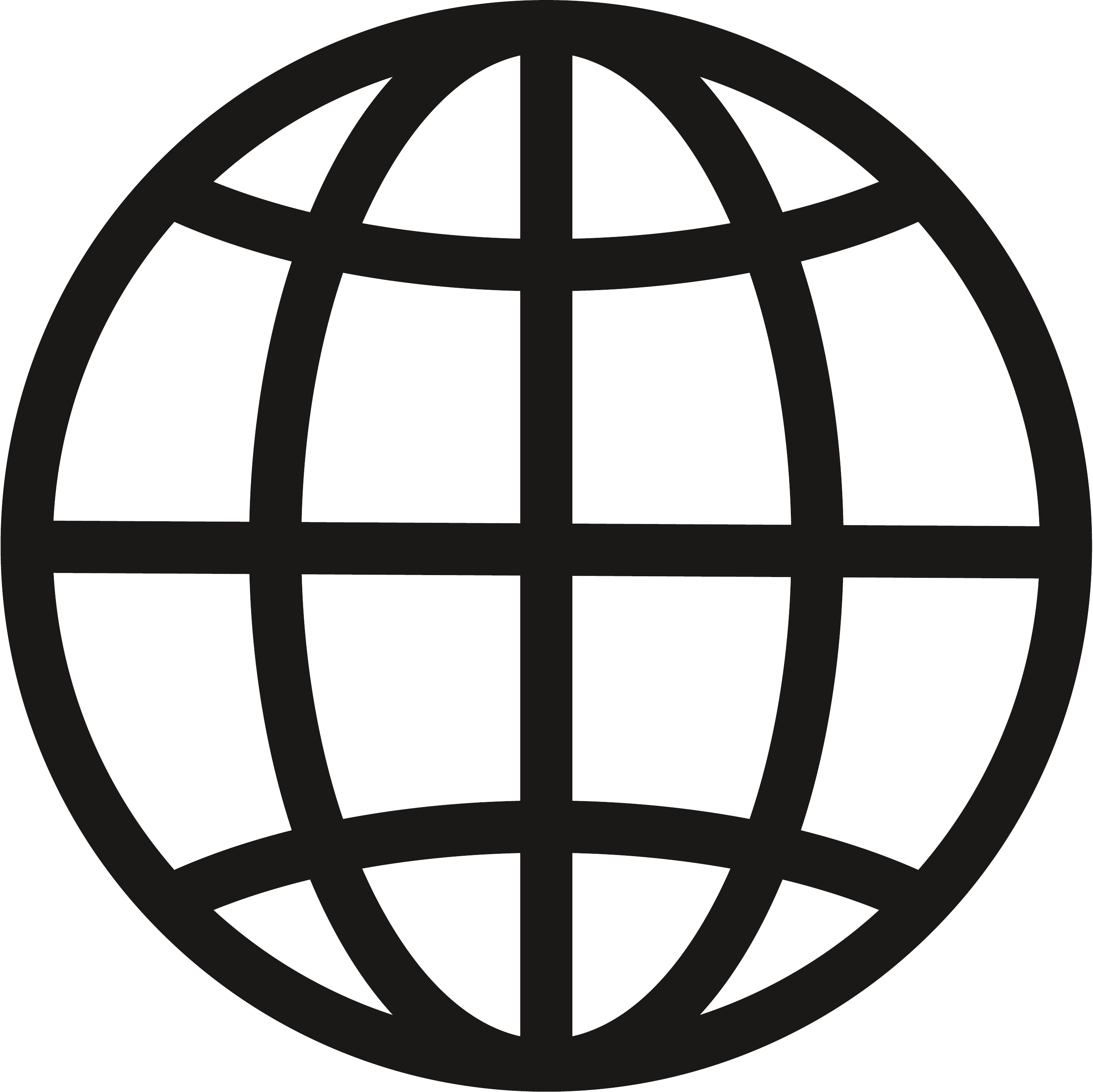}}\xspace}

\definecolor{scholarblue}{rgb}{0.21,0.49,0.74}
\definecolor{bluelink}{RGB}{0,113,188}
\definecolor{greenlink}{RGB}{0,188,113}
\hypersetup{
    colorlinks=true,%
    citecolor=scholarblue,%
    filecolor=red,%
    linkcolor=red!93!black,%
    urlcolor=bluelink
}

\usepackage{titlesec}
\titlespacing*{\paragraph}
    {0pt}     %
    {0.25em}  %
    {1em}  %

\definecolor{navyblue}{HTML}{0071BC}

\newcommand{\ourscore}[1]{Multi-Agent World Score}
\newcommand{\oursf}{Checkpointed Self Forcing\xspace}
\newcommand{\matrixgame}{Matrix Game 2.0\xspace}

\newcommand\solaris{{\fontfamily{lmtt}\selectfont Solaris}\xspace}
\newcommand\solarisengine{{\fontfamily{lmtt}\selectfont SolarisEngine}\xspace}

\definecolor{diffgreen}{RGB}{213, 245, 213}
\definecolor{diffred}{RGB}{255, 215, 215}

\definecolor{diffgreen}{RGB}{213, 245, 213}
\definecolor{diffred}{RGB}{255, 215, 215}

\newcommand{\added}[1]{%
  {\setlength{\fboxsep}{1.5pt}\colorbox{diffgreen}{\texttt{+} #1}}%
}
\newcommand{\removed}[1]{%
  {\setlength{\fboxsep}{1.5pt}\colorbox{diffred}{\texttt{-} #1}}%
}

\definecolor{blindcolor}{HTML}{AB2AC6}    %
\definecolor{chancecolor}{HTML}{F59E0B}   %
\definecolor{singlecolor}{HTML}{06B6D4}   %
\definecolor{multiplecolor}{HTML}{2563EB} %
\definecolor{captioncolor}{HTML}{22C55E}  %

 \newcommand{\culine}[2]{%
    \def\temp@uline{\bgroup\markoverwith
        {\textcolor{#1}{\rule[-0.5ex]{2pt}{1pt}}}\ULon}%
    \temp@uline{#2}%
}
 \newcommand{\cthickuline}[3][0.8pt]{%
    \def\temp@uline{\bgroup\markoverwith
        {\textcolor{#2}{\rule[-0.5ex]{2pt}{#1}}}\ULon}%
    \temp@uline{#3}%
}

\title{\center{\solaris: \\ Building a Multiplayer Video World Model in Minecraft}}

\renewcommand\Affilfont{\normalfont\fontsize{11}{15}\selectfont\centering}
\author{
    Georgy~Savva$^{\dagger *}$\quad
    Oscar~Michel$^{\dagger *}$\quad
    Daohan~Lu$^{*}$\quad
    Suppakit~Waiwitlikhit\quad \\
    Timothy~Meehan\quad 
    Dhairya~Mishra\quad 
    Srivats~Poddar\quad
    Jack~Lu\quad 
    Saining~Xie \\
    New York University
}

\begingroup

\footnotetext{$^\dagger$ Project lead; order determined by coin toss.}
\footnotetext{$^*$ Equal technical contribution.}
\endgroup

\begin{abstract}
Existing action-conditioned video generation models (video world models) are limited to single-agent perspectives, failing to capture the multi-agent interactions of real-world environments. We introduce \solaris, a multiplayer video world model that simulates consistent multi-view observations. To enable this, we develop a multiplayer data system designed for robust, continuous, and automated data collection on video games such as Minecraft. Unlike prior platforms built for single-player settings, our system supports coordinated multi-agent interaction and synchronized videos + actions capture. Using this system, we collect 12.64 million multiplayer frames and propose an evaluation framework for multiplayer movement, memory, grounding, building, and view consistency. We train \solaris using a staged pipeline that progressively transitions from single-player to multiplayer modeling, combining bidirectional, causal, and Self Forcing training. In the final stage, we introduce \emph{\oursf}, a memory-efficient Self Forcing variant that enables a longer-horizon teacher. Results show our architecture and training design outperform existing baselines. Through open-sourcing our system and models, we hope to lay the groundwork for a new generation of multi-agent world models.
\end{abstract}

\begin{document}

\maketitle

\begin{center}
    \renewcommand{\arraystretch}{1.5}
    \begin{tabular}{rll}
        \worldwideweb{} & \textbf{Website} & \url{https://solaris-wm.github.io/}\\
        \github{} & \textbf{Engine Code} & \url{https://github.com/solaris-wm/solaris-engine}\\
        \github{} & \textbf{Model Code} & \url{https://github.com/solaris-wm/solaris}\\
        \huggingface{} & \textbf{Datasets} & \url{https://huggingface.co/collections/nyu-visionx/solaris-data} \\
        \huggingface{} & \textbf{Models} & \url{https://huggingface.co/collections/nyu-visionx/solaris-models} \\

    \end{tabular}
\end{center}

\newpage
{
    \hypersetup{linkcolor=black}
    \tableofcontents
}
\newpage

\section{Introduction}
\label{sec:intro}
\begin{figure*}[h!]
    \centering
    \includegraphics[width=0.95\linewidth]{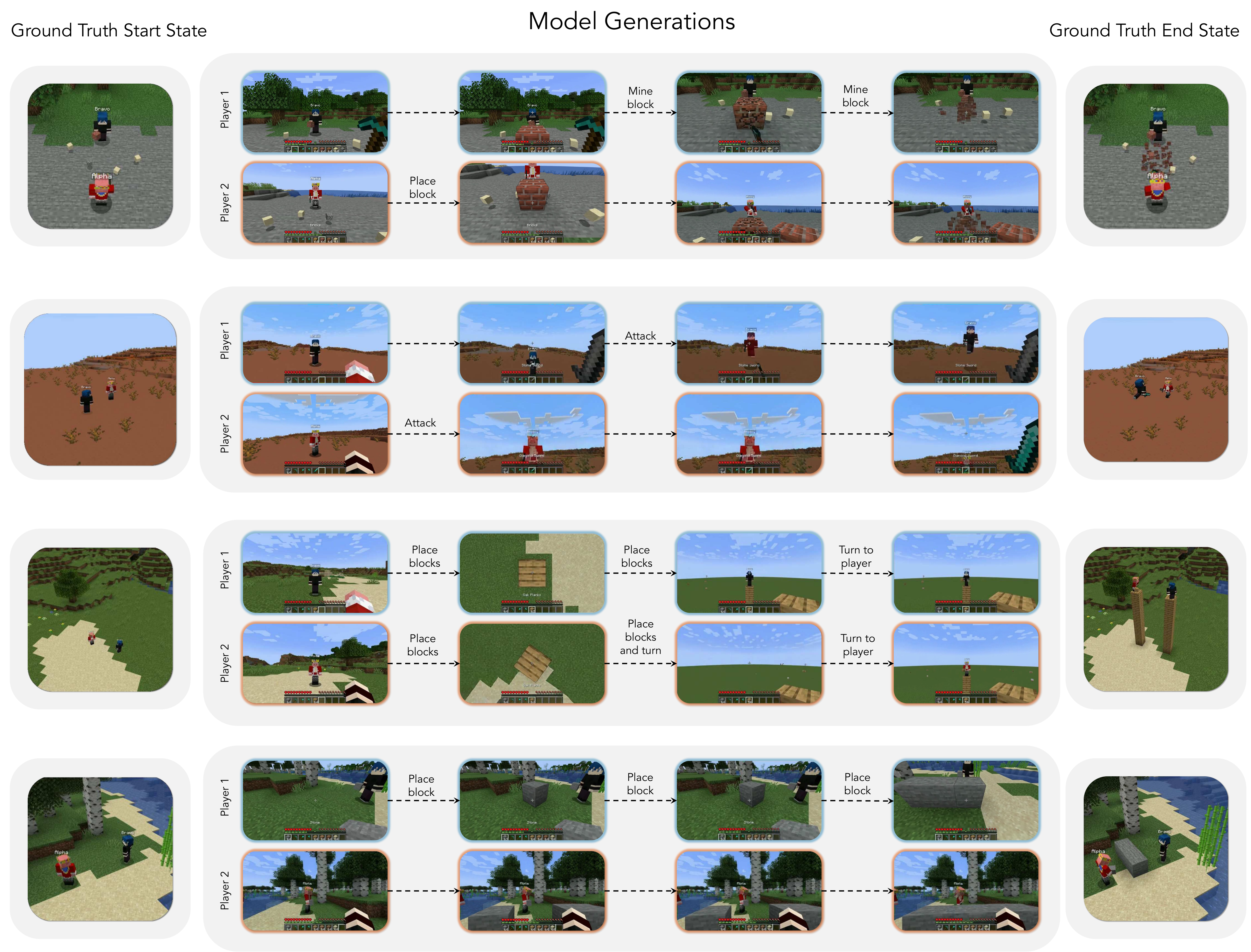}
    \caption{\textbf{Selected samples from our model.} Our model takes in starting frames from each player as input and generates action-conditioned videos. The action descriptions shown here are summaries of the fine-grained action sequences given to the model that span many frames. The third-person ground truth visualizations are not given to the model.}
    \label{fig:teaser}
\end{figure*}

Video world models, where future agent observations are generated based on past observations and actions, hold tremendous promise as a tool for embodied AI agents, and are useful for synthetic training~\cite{ali2025world, yang2024video}, inference-time planning~\cite{zhou2024dino, assran2025v}, and policy learning and evaluation~\cite{du2023learning, jang2025dreamgen, quevedo2026worldgym}. However, today’s world models are limited to simulating the observations of only a single agent at a time. In our fundamentally multi-agent world, video world models that are capable of simulating the perspectives of all agents in an environment are needed to capture an accurate world state. 

Toward this end, we study the creation of multiplayer video world models trained on multiplayer data collected from Minecraft. Modeling the perspectives of multiple players at once is significantly more challenging than single-player modeling. Generated observations need to be consistent not only across time but also across agent perspectives. Actions taken by one agent, such as movement or block placement, must be simultaneously and accurately reflected in the viewpoints of all other agents.

Minecraft serves as an ideal testbed for multi-agent modeling. Its unbounded, fully 3D worlds rigorously test challenging aspects of viewpoint modeling, such as perspective consistency, occlusion handling, and spatial memory. Its dynamic and malleable environment tests a model’s ability to keep track of alterations that occur over time. Furthermore, environmental stochasticity (e.g., mobs, weather) forces models to disentangle alterations caused by environmental randomness from those caused by controllable agents. And through its complex system of building, digging, and crafting, Minecraft allows for endless complexity. Lastly, its procedurally generated terrain naturally diversifies the environment. We believe Minecraft will continue to be a valuable research platform in the long-term.

Prior to this work, there were no existing publicly available systems for simulating multiplayer Minecraft gameplay, motivating us to build \solarisengine: a system for collecting gameplay from pre-programmed bots. In \solarisengine, our bots engage in non-trivial and realistic gameplay, including mining, attacking, building, and navigating terrains. Our system allows us to collect millions of frames in a matter of hours, and its modular design makes it easy to extend to new agent behaviors. We used this data engine to generate a multiplayer training dataset of 12.64 M frames (6.32 M per player) and an evaluation set testing multiplayer movement, memory, grounding, building, and view consistency.

Our model, called \solaris, is a video diffusion model with a general-purpose architecture that adapts a single-agent pre-trained video DiT to one that simulates multiple perspectives with minimal modifications. Our model takes in sequences of past player observations and actions and generates future observations using autoregressive diffusion. We show examples of our model's capabilities in~\cref{fig:teaser}.

To improve long-horizon autoregressive generations, we adopt the Self Forcing~\cite{huang2025selfforcingbridgingtraintest} paradigm. We extend the setting of Self Forcing to allow the student to benefit from a long-context teacher. Naively applying Self Forcing with sliding-window generation leads to excessive memory usage, a problem we mitigate with a novel technique called \oursf analogous to gradient checkpointing. We also make simplifications to the initialization of the student model, finding that the initialization method of CausVid~\cite{yin2025slowbidirectionalfastautoregressive} can be replaced by simply finetuning with causal masking. 

In summary, we present \solaris, a multi-agent video world model capable of simulating consistent multiplayer perspectives in Minecraft. We contribute a scalable data system, \solarisengine, a large-scale training dataset of multiplayer gameplay, an evaluation system, and a new architecture that adapts a pre-trained video DiT for multiplayer videos. Finally, we introduce \oursf, a memory-efficient extension of the Self Forcing paradigm, enabling effective autoregressive training.

\section{Related Work}
\label{sec:related_work}

\subsection{World Models and Video World Models} 
A world model takes in an observation of a system along with a chosen action or intervention, and produces a prediction of the resulting state or outcome. The notion of a world model can be traced back to Kenneth Craik, who argued in his 1943 book \textit{The Nature of Explanation}~\cite{craik1943nature} that if an organism carries a small internal model of reality and its possible actions, it can test options mentally, anticipate the future, use past experience, and respond more effectively and safely to challenges. These ideas later became central to dynamic systems and control theory~\cite{bellman1957dynamic, kalman1960new, bryson1975applied}, establishing mathematical tools for modeling, prediction, and planning in uncertain environments. The Dyna architecture proposed by Sutton~\cite{sutton_dyna1991} emphasized that learning an internal model of the world enables agents to plan their actions rather than rely solely on reactive trial-and-error interactions with the environment, such as in model-free reinforcement learning. More recently, the concept of world models has been combined with deep generative modeling. Several works ~\cite{ha2018world, hafner2020dreamcontrollearningbehaviors, hafner2020mastering, hafner2023mastering, hafner2025trainingagentsinsidescalable} demonstrated learning a compact latent dynamics model directly from pixel observations, enabling policy optimization within a learned latent environment. Instead of modeling raw pixels or compression-based latent spaces such as those from variational autoencoders~\cite{kingma2013auto} that remain close to pixel space~\cite{zheng2025diffusion}, an important direction is for world models to learn abstract representations where predictions can ignore irrelevant details, enabling better understanding, prediction, and planning~\cite{zhou2024dino, bardes2024revisiting, assran2025v}.

With the emergence of video diffusion transformers~\cite{peebles2023scalablediffusionmodelstransformers} and major advances in text-to-video and image-to-video generation, a growing trend in world modeling is to use video diffusion models as \emph{world simulators}~\cite{openai2024sora,yang2023unisim, liang2024dreamitate, bar2025navigation}. Since this line of work typically does not focus on learning abstract representations or on planning and decision making within them, we refer to these approaches as \emph{video world models} to emphasize their reliance on video diffusion for simulation. Video world models have been applied to numerous domains including robotics~\cite{mereu2025generative, yang2024learninginteractiverealworldsimulators, li2025unified,li2026causal,ye2026world,gao2026dreamdojo}, video games~\cite{oasis2024, parkerholder_fruchter_2025_genie3, valevski2025diffusionmodelsrealtimegame, alonso2024diffusionworldmodelingvisual, he2025matrixgame20opensourcerealtime, xiao2025worldmem}, self-driving~\cite{agarwal2025cosmos,hu2023gaia,waymo2026wm}, and physical simulation~\cite{li2025pisa, yuan2026inference}. Our work contributes to this direction. However, our focus is on building a \emph{multiplayer} video world model. Instead of just generating pixels from actions, we want to create a foundation for agent learning in a setting where multiple agents share the same world, and where simply simulating pixels is not enough. To our knowledge, Multiverse~\cite{enigma2025multiverse} is the only video world model capable of simulating multiple agents. While their U-Net~\cite{ronneberger2015u} model is trained on gameplay from one race track (``Tsukuba'') in the 2004 video game Gran Turismo 4, we tackle the significantly more complex 3D open-world environment of Minecraft. We study applying their channel-concatenation design to our model in~\cref{sec:experiments}.

\subsection{Autoregressive Video Generation}

Large scale Diffusion Transformer (DiT)~\cite{peebles2023scalablediffusionmodelstransformers} video generation models have made tremendous progress in recent years~\cite{openai2024sora, kong2025hunyuanvideosystematicframeworklarge, wan2025wanopenadvancedlargescale}. Diffusion Forcing~\cite{chen2024diffusionforcingnexttokenprediction} is a technique where autoregressive generation emerges as a byproduct of training with an independent noise level per frame. CausVid~\cite{yin2025slowbidirectionalfastautoregressive} was introduced to convert a bidirectional video model to an efficient causal model of comparable quality. Self-Forcing~\cite{huang2025selfforcingbridgingtraintest,cui2025self, hong2025relic} improves generation quality further by supervising on a model's own generations, mitigating autoregression train-test mismatch.  

A concurrent work to ours, RELIC~\cite{hong2025relic}, also studies Self-Forcing with a long-context teacher. Like our method, they propose a memory-efficient implementation involving a recomputation step for the backward pass. However, we distinguish our approach by performing this step in one parallel forward pass via masking, avoiding the multiple rolling passes required by RELIC.

\subsection{AI Agents in Minecraft}

A number of platforms for agents in Minecraft have been developed, which have primarily been used for research in reinforcement learning~\cite{johnson2016malmo, fan2022minedojo, guss2019minerllargescaledatasetminecraft}. Mineflayer~\cite{mineflayer2025} is a commonly used framework for developing bots in Minecraft, and was used to collect the LoopNav~\cite{lian2025toward} dataset testing spatial memory, as well as in Voyager~\cite{wang2023voyager}, a Minecraft LLM agent. VPT~\cite{baker2022videopretrainingvptlearning} is a large-scale dataset of single-player Minecraft data collected from humans. However, as we explain in~\cref{sec:system}, none of these frameworks are capable of simulating multiplayer gameplay with visuals, leading us to develop \solarisengine. 

\begin{figure}[h]
    \centering    \includegraphics[width=\linewidth]{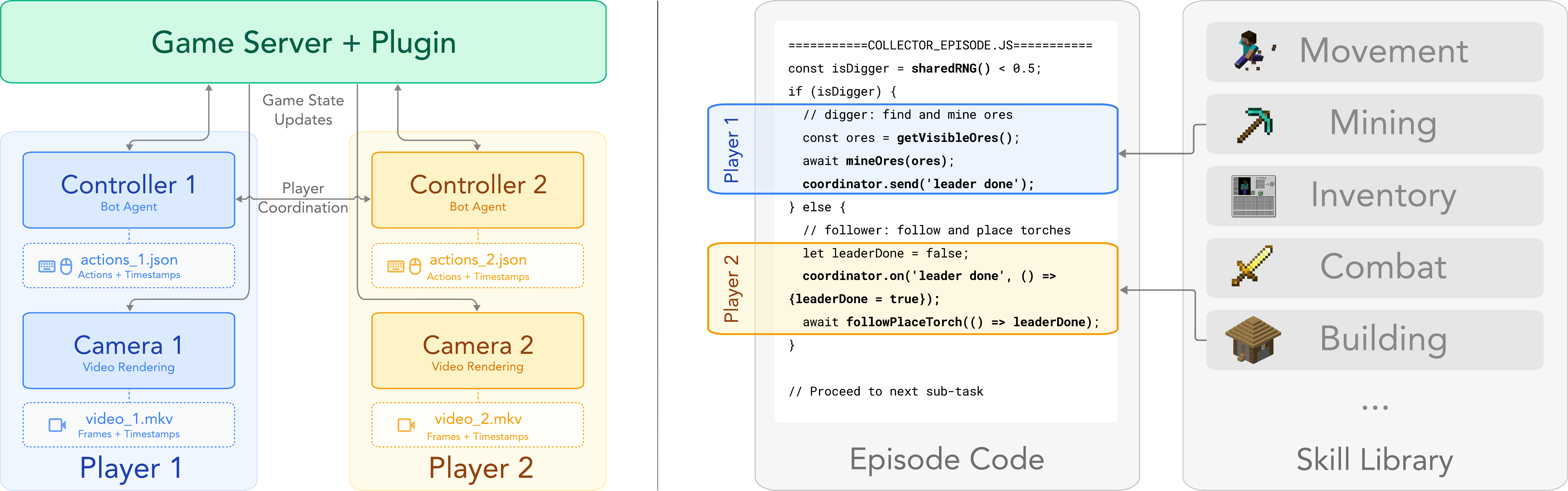}
    \caption{\textbf{\solarisengine Overview.} \emph{(Left)} Docker-based orchestration of containerized game server, camera, and controller bots. Cameras mirror Controllers' state and actions via a custom server-side plugin; Controllers are Mineflayer bots that run episode code and log low-level actions. \emph{(Right)} Episodes compose reusable skill primitives from a shared library. Simplified ``collector'' episode code is shown.}
    \label{fig:solaris-engine}
\end{figure}

\section{\solarisengine: A Framework for Multiplayer Gameplay at Scale}
\label{sec:system}
\begin{table}[t]
\centering

\begin{tabular}{lccc}
\toprule
\textbf{Method} & \textbf{Controllability} & \textbf{Multiplayer} & \textbf{Graphics} \\
\midrule
Malmo~\cite{johnson2016malmo} & \texttimes & \checkmark & \checkmark \\
MineRL~\cite{guss2019minerllargescaledatasetminecraft}     & \texttimes & \texttimes & \checkmark \\
MineDojo~\cite{fan2022minedojo}  & \texttimes & \texttimes & \checkmark \\
Voyager~\cite{wang2023voyager} & \checkmark & \texttimes & \texttimes \\
Mineflayer~\cite{mineflayer2025} & \checkmark & \checkmark & \texttimes \\
\solarisengine    & \checkmark & \checkmark & \checkmark \\
\bottomrule
\end{tabular}
\caption{\textbf{Comparison of Minecraft AI frameworks.} Unlike prior systems, \solarisengine enables controlled multiplayer gameplay collection with real Minecraft graphics. RL-based frameworks such as Malmo, MineRL, and MineDojo produce visual observations but offer limited controllability, as their low-level action spaces require training RL agents to collect meaningful data, which is prohibitively expensive and counterproductive for world-modeling data collection. Voyager and Mineflayer provide high-level behavioral control but operate in text-only mode without visual output. \solarisengine supports multiplayer gameplay, fine-grained programmatic control and rich visual observations.}
\label{tab:mc_systems_comparison}
\end{table}

We describe our core system, \solarisengine, built from the ground up for capturing pre-programmed multiplayer Minecraft gameplay and large-scale data collection. Several frameworks exist for controlling agents in Minecraft, including Malmo~\cite{johnson2016malmo}, MineRL~\cite{guss2019minerllargescaledatasetminecraft}, Minedojo~\cite{fan2022minedojo}, and Mineflayer~\cite{mineflayer2025}. While these tools offer various capabilities (see~\cref{tab:mc_systems_comparison}), we found they could not be adapted to generate coherent, cooperative multiplayer gameplay. Our system enables us to collect millions of frames through a primitive skills library, a multiplayer communication layer, and a modular, extensible episode system. Looking ahead, we envision this foundation serving other purposes, such as collecting data for Vision-Language-Action models or testing large-scale collaboration among AI agents.

\begin{figure*}[h!]
    \centering
    \includegraphics[width=\linewidth]{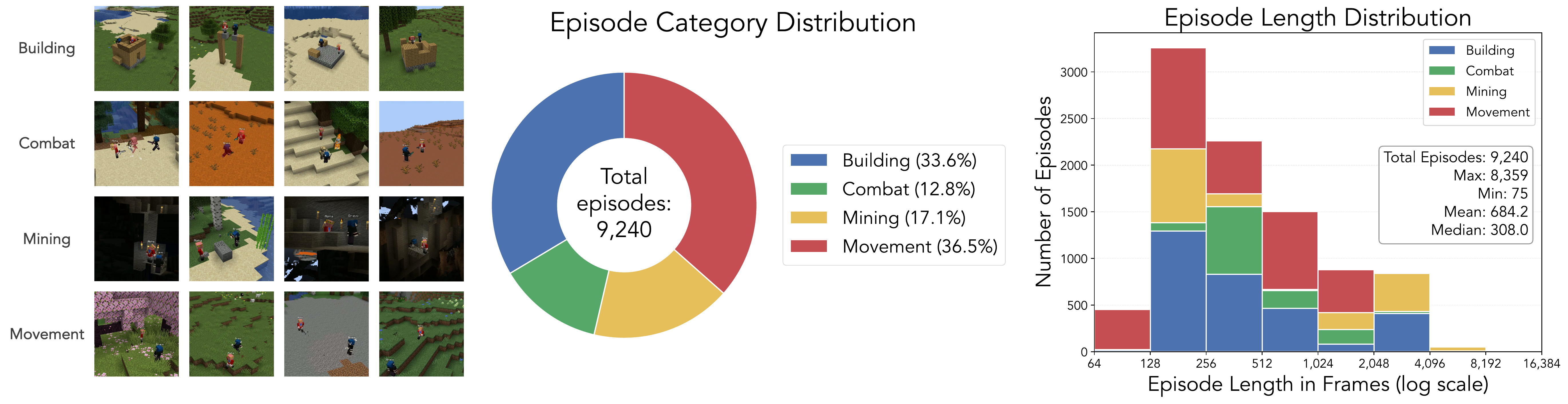}
    \caption{\textbf{Dataset Statistics of our training dataset.} \emph{(Left)} The dataset consists of four different episode categories focusing on building, combat, movement, and mining scenarios, respectively. \emph{(Middle)} It has a total of 9,240 episodes and 6.32~M frames per player, for a combined 12.64~M frames. Episode types are chosen randomly with weights that decrease with respect to the typical episode length. \emph{(Right)} Most episode lengths range from 128 to 512 frames or 6.4 to 25.6 seconds (we record at 20 fps).}
    \label{fig:dataset-plots}
\end{figure*}

\subsection{Enabling cooperative multiplayer gameplay} Existing frameworks like Malmo~\cite{johnson2016malmo}, MineRL~\cite{guss2019minerllargescaledatasetminecraft}, and Minedojo~\cite{fan2022minedojo} use low-level action spaces, where, without RL training, agents can only take random actions. Data from these agents would be too simple and chaotic for world modeling, a problem that gets worse with multiple agents. Though agents could be trained using RL to learn meaningful action policies, the visual data collected from such agents would be geared toward collecting rewards and would not necessarily be diverse, realistic, and human-like.

Data for training video world models, on the other hand, demands realistic gameplay. We use Mineflayer, a JavaScript Minecraft client library that the community has used for building game bots. Notably, in Voyager~\cite{wang2023voyager}, it is also used as a high-level API for programming agents, albeit in a single-player setting. Mineflayer provides primitives like \textit{pathfinding}, \textit{block placement}, and \textit{combat} that are composable, letting us generate quality gameplay through programming alone.

However, Mineflayer was not designed for multiplayer coordination. To address Mineflayer's lack of multiplayer support, we built a communication layer enabling bot coordination to mimic collaborative human gameplay. We introduced high-level primitives such as building, scaffolding, tool use, and navigation. These primitives, shown in~\cref{fig:solaris-engine}, when combined with the communication layer, form an episode where two bots achieve a predefined goal. We created a library of these episode types covering core Minecraft interaction aspects. For a full list, please refer to~\cref{subsec:appendix_episode_types}.

Although our episode library is written in high-level code using predefined primitives, the system translates these into low-level actions as if they were collected from a human player, making our dataset compatible with VPT \cite{baker2022videopretrainingvptlearning}. See ~\cref{subsec:appendix_action_space} for the description of all supported actions and the relation to VPT.

\subsection{Extracting and aligning visuals with actions}

Mineflayer has another significant limitation: it lacks rendering capabilities. To record visuals, we pair each controller bot with a "camera bot" running the official Minecraft Java client in headless mode with GPU-accelerated rendering. A custom server-side plugin synchronizes the camera bot to mirror the controller's state and actions, even including animations, in real time. Although implemented as separate processes, they form a single logical player (see~\cref{fig:solaris-engine}), with actions and graphical observations aligned via timestamps in a post-processing step. While we currently run with two, the system architecture can theoretically accommodate any number of concurrent players.

\subsection{Robust and Scalable Docker System}
We implement the controller bots, camera bots, and Minecraft server as Docker containers, orchestrated through Docker Compose to run in isolated units. A suite of Python scripts manages these units, launching multiple Compose workers in parallel to enable scalable data collection. The bots operate in a loop, sampling and executing episodes from our library. The bots teleport to a random location at the start of an episode to diversify terrain.

Minecraft's complexity and stochasticity mean that episodes inevitably encounter errors or get stuck. To handle this, we built a safety mechanism that detects episode failures during execution, notifies all components, and aborts the current episode across all bots. The system then proceeds to a new episode with a fresh state, ensuring continuous data collection without manual intervention.

\begin{figure*}[h!]
    \centering
    \includegraphics[width=\linewidth]{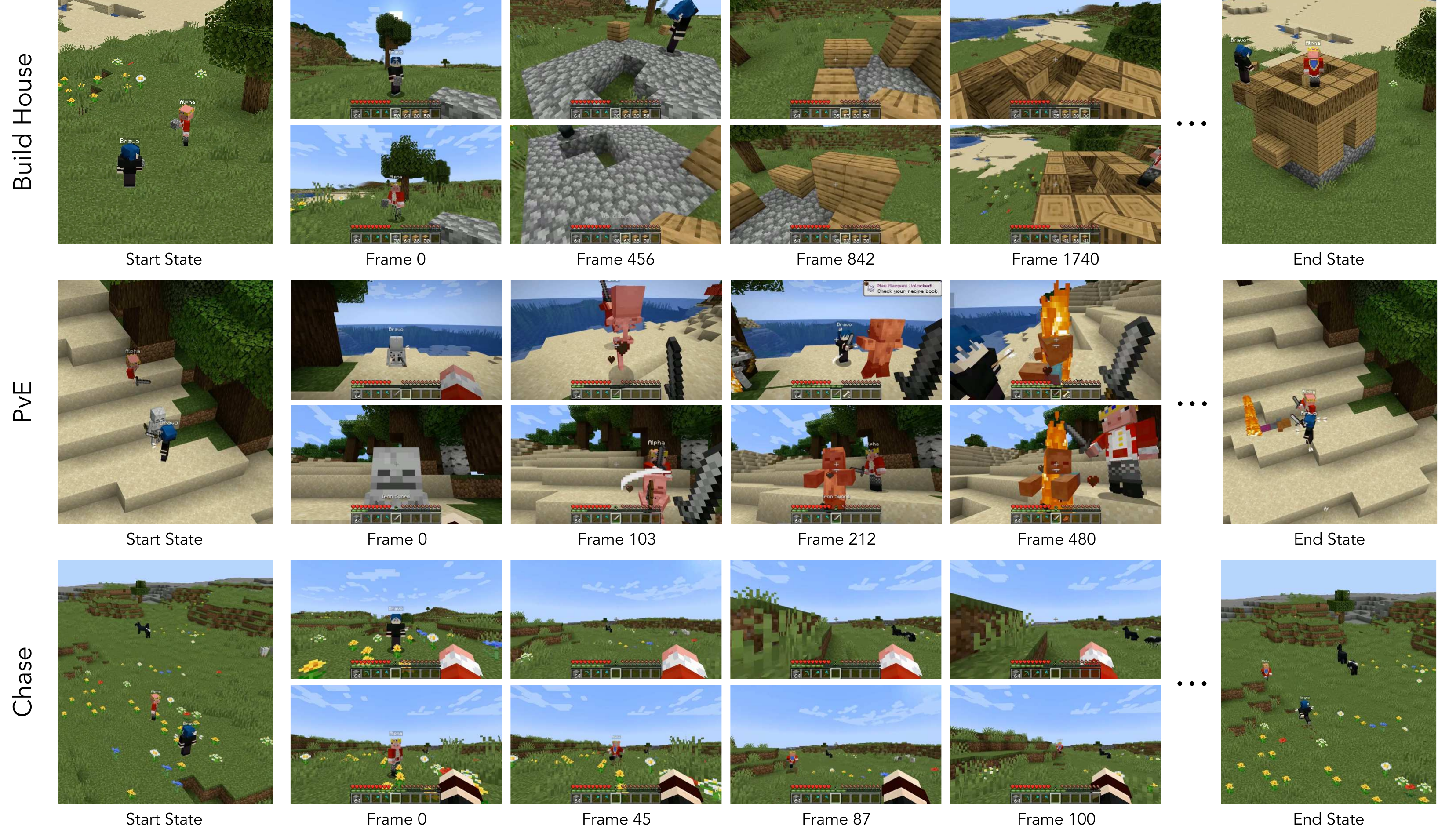}
    \caption{\textbf{Episode Demonstrations from our training dataset.} We show the recorded frames from 3 different training episodes at various points in time. Note that the third-person ``start state'' and ``end state'' screenshots are for visualization only and are not part of the dataset.}
    \label{fig:dataset-demo}
\end{figure*}

\subsection{Multiplayer Training Dataset}

Using our framework, we collect a large-scale multiplayer training dataset. The dataset episodes, shown in~\cref{fig:dataset-demo}, cover core aspects of the game: building, combat, movement, and mining. Our bots build houses, walls, towers, and bridges, fight mobs or each other, chase each other and navigate together, dig to find valuable ores, mine their way underground towards each other, or mine blocks on the ground. We collect our episodes in an even split between ``Superflat'' and ``Normal'' Minecraft world types. To the best of our knowledge, our dataset is the first action-annotated multiplayer Minecraft dataset suitable for machine learning applications.

The gameplay features diverse times of day, biomes, in-game tool use, and weather conditions. Our action annotations span a wide range of Minecraft environment actions: WASD movement, jumping, sprinting, sneaking, camera changes, and interactive actions such as digging, placing, attacking, and item switching (see ~\cref{subsec:appendix_action_space} for the full action space). We collect a total of $6.32$ M frames per player. We show important dataset statistics in~\cref{fig:dataset-plots}.

\begin{figure}[h!]
    \centering
    \includegraphics[width=0.5\linewidth]{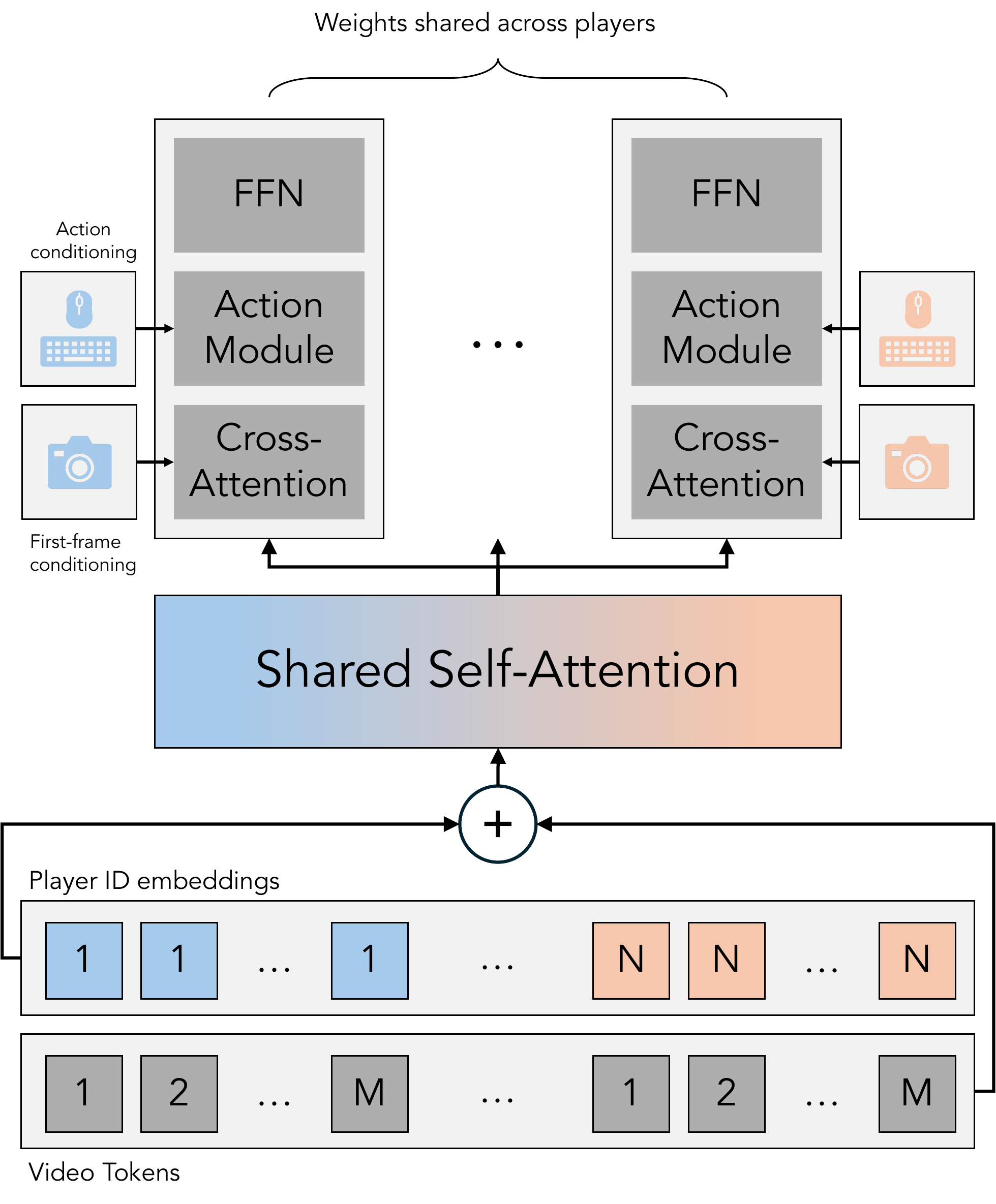}
    \caption{Our modified DiT block achieves multiplayer modeling through visual interleaving along the sequence dimension. We denote the number of players with $N$ and the number of tokens per video with $M$. Multiplayer information is exchanged through a shared self-attention block. The other modules are unchanged from \matrixgame and applied independently per player.}
    \label{fig:arch}
\end{figure}

\section{\solaris Model Design}
\label{sec:solaris_model}

\subsection{Preliminaries}
We consider a world model that can predict the future observations of multiple agents given their past observations and actions. While a traditional autoregressive diffusion model operates on a latent frame $\mathbf{x}_{\text{SP}} \in \mathbb{R}^{H \times W \times C}$, we generalize this to the multi-agent setting by augmenting the state space to include an extra player dimension $\texttt{P}$, performing diffusion on a joint tensor of shape $\texttt{(P, H, W, C)}$. Let $\mathbf{x}^t = \{x_1^t, \dots, x_P^t\}$ denote the joint state and $\mathbf{a}^{t} = \{a_1^t, \dots, a_P^t\}$ the joint actions of all $P$ agents at time $t$. Let $\mathbf{x} \coloneqq \mathbf{x}^{1:T}$ and $\mathbf{a} \coloneqq \mathbf{a}^{1:T}$ denote the combined tensor of consecutive latent observation and action tensors across a sequence of length $T$. Thus $\mathbf{x}$ has shape \texttt{(B, P, T, H, W, C)} and $\mathbf{a}$ has shape \texttt{(B, P, T, D)}. We model the probability distribution $p_\theta(\mathbf{x}) = \prod_{t=1}^{T} p_\theta(\mathbf{x}^t \mid \mathbf{x}^{<t}, \mathbf{a}^{<t})$ using a diffusion model. We restrict $P=2$ in this work, though our framework is flexible enough to be generalizable to any number of players.

We train our model with conditional Flow Matching~\cite{liu2022flow,lipman2022flow}. Concretely, we optimize the loss
\[
\mathcal{L}_\theta = \mathbb{E}_{\mathbf{x}, \mathbf{a},\sigma, \epsilon} \left[ \left\| v_\theta(\mathbf{x}_{\boldsymbol{\sigma}}, \boldsymbol{\sigma}, \mathbf{a}) - (\epsilon - \mathbf{x}) \right\|_2^2 \right],
\]
using the standard forward process $\mathbf{x}_{\boldsymbol{\sigma}} = (1 - \boldsymbol{\sigma})\mathbf{x} + \boldsymbol{\sigma}\epsilon$ where $\epsilon\sim \mathcal{N}(\mathbf{0}, \mathbf{I})$. The noise schedule differs depending on the model variant. For our \emph{bidirectional} model, we use a shared noise level across all players and frames: $\sigma \sim \mathcal{U}(0, 1)$, with $\boldsymbol{\sigma} = \sigma \cdot \mathbf{1}_{ P \times T}$, so that all elements are noised and denoised jointly. For our \emph{causal} model, we adopt Diffusion Forcing~\cite{chen2024diffusionforcingnexttokenprediction} to enable autoregressive generation, sampling independent noise levels per player and per frame: $\boldsymbol{\sigma} \in [0, 1]^{P \times T}$, where each entry $\sigma_{p,t} \sim \mathcal{U}(0, 1)$ is independent.

\subsection{Network Architecture}
\label{subsec:model_architecture}

Our architecture is built on \matrixgame~\cite{he2025matrixgame20opensourcerealtime}, a single-player controllable video Diffusion Transformer~\cite{peebles2023scalablediffusionmodelstransformers} (DiT) model that was trained on multiple video games, including Minecraft. A visualization of our multiplayer DiT block can be found in~\cref{fig:arch}. To adapt \matrixgame to be capable of multiplayer modeling, we introduce several changes which we discuss below. 

\textbf{Expanded action space.} We extend the action space of \matrixgame to use the full range of Minecraft actions encoded as the MineRL~\cite{guss2019minerllargescaledatasetminecraft} action space by increasing the input dimension of the \matrixgame keyboard action module and reinitializing its weights. We run the action module independently per-player, folding the player dimension into the batch. In psuedocode, this is achieved through \texttt{rearrange("B P T D -> (B P) T D")}.  

\textbf{Multiplayer attention.} Information across different players is exchanged through Multiplayer Self-Attention layers in our DiT blocks. We apply 3D RoPE~\cite{su2024roformer} to each player's tokens \textit{independently} and inject player information by adding learned player ID embeddings to each player's tokens at the start of each Multiplayer Self-Attention layer. The cross-attention block, which provides first-frame conditioning, remains unchanged from \matrixgame and is applied independently per-player. 

\begin{figure}[h!]
    \centering
    \includegraphics[width=0.6\linewidth]{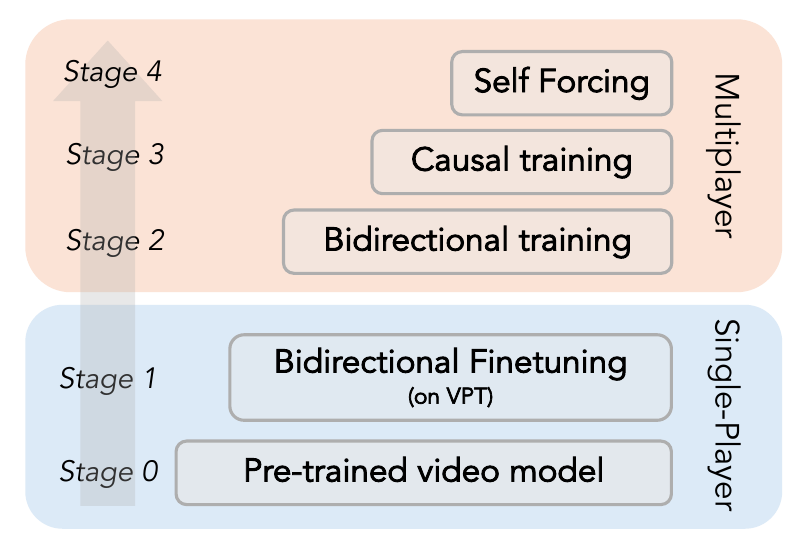}
    \caption{\textbf{An overview of the full training pipeline.} Starting with a pretrained bidirectional video diffusion model, we first finetune it with single-player and then multiplayer data. We then finetune it with a causal mask before using Self Forcing to achieve stable long-horizon autoregressive generations.}
    \label{fig:pipeline}
\end{figure}

\section{Multiplayer Training Pipeline}
\label{sec:model_architecture}
We follow the established practice of first training a bidirectional base model and then adapting it to be causal, enabling auto-regressive generation. We illustrate the stages of our training process in~\cref{fig:pipeline}.

\subsection{Stage 1: Bidirectional Single-Player}
\label{sec:bidirectional_sp_training}

We begin by initializing our model from the weights\footnote{Specifically, we use the \href{https://huggingface.co/Skywork/Matrix-Game-2.0/tree/main/base_distilled_model}{\texttt{base\_distilled\_model}} checkpoint.} of Matrix-Game 2.0~\cite{he2025matrixgame20opensourcerealtime}. \matrixgame is trained to support multiple video games in addition to Minecraft and as a result has a limited action space consisting solely of viewpoint (camera) and directional (WASD) movement. We adapt our model to support the full range of Minecraft actions by finetuning it on the VPT~\cite{baker2022videopretrainingvptlearning} dataset, which contains over 2,000 hours of human gameplay. We train the model with bidirectional attention for 120K steps with a context length of 33 frames. We find this single-player pretraining stage to be highly important as it gives our model an effective initialization for multiplayer modeling. We study this experimentally in~\cref{sec:experiments}. Throughout all stages of training, we  use the 3D VAE from \matrixgame and keep it frozen.

\subsection{Stage 2: Bidirectional Multiplayer} 

We next adapt our single-player model to support multiplayer data. We employ the architectural modifications described in~\cref{subsec:model_architecture} and train the model with full-sequence diffusion on our dataset for 120k steps, where we find that FID on a held-out test set converges at this point. This checkpoint is used as the teacher during Self Forcing (see~\cref{subsec:self-forcing} for details). 

\subsection{Stage 3: Causal Multiplayer}

To save training time, we branch training once the bidirectional model reaches 60k steps. We use this intermediate checkpoint to initialize the causal model, while the bidirectional model continues to train independently to 120k steps to serve as a high-quality teacher. Following \matrixgame, the causal model uses a sliding window attention mask with a window size of 6 latent frames (24 real frames), which serves as the maximum size of the rolling KV cache during inference. We train this model for 60k steps using diffusion forcing, and it serves as the initialization for the generator in Self Forcing. 

Our strategy of using Diffusion Forcing with a causal mask to initialize the generator is simpler than the standard CausVid~\cite{yin2025slowbidirectionalfastautoregressive} initialization of Self Forcing, which involves ODE regression and few-step distillation with DMD~\cite{yin2024one, yin2024improved} \textit{before} Self Forcing. We find our approach compares favorably to CausVid. We study this experimentally; please see~\cref{subsec:self-forcing} for details. 

\begin{figure*}[h!]
    \centering
    \includegraphics[width=\linewidth]{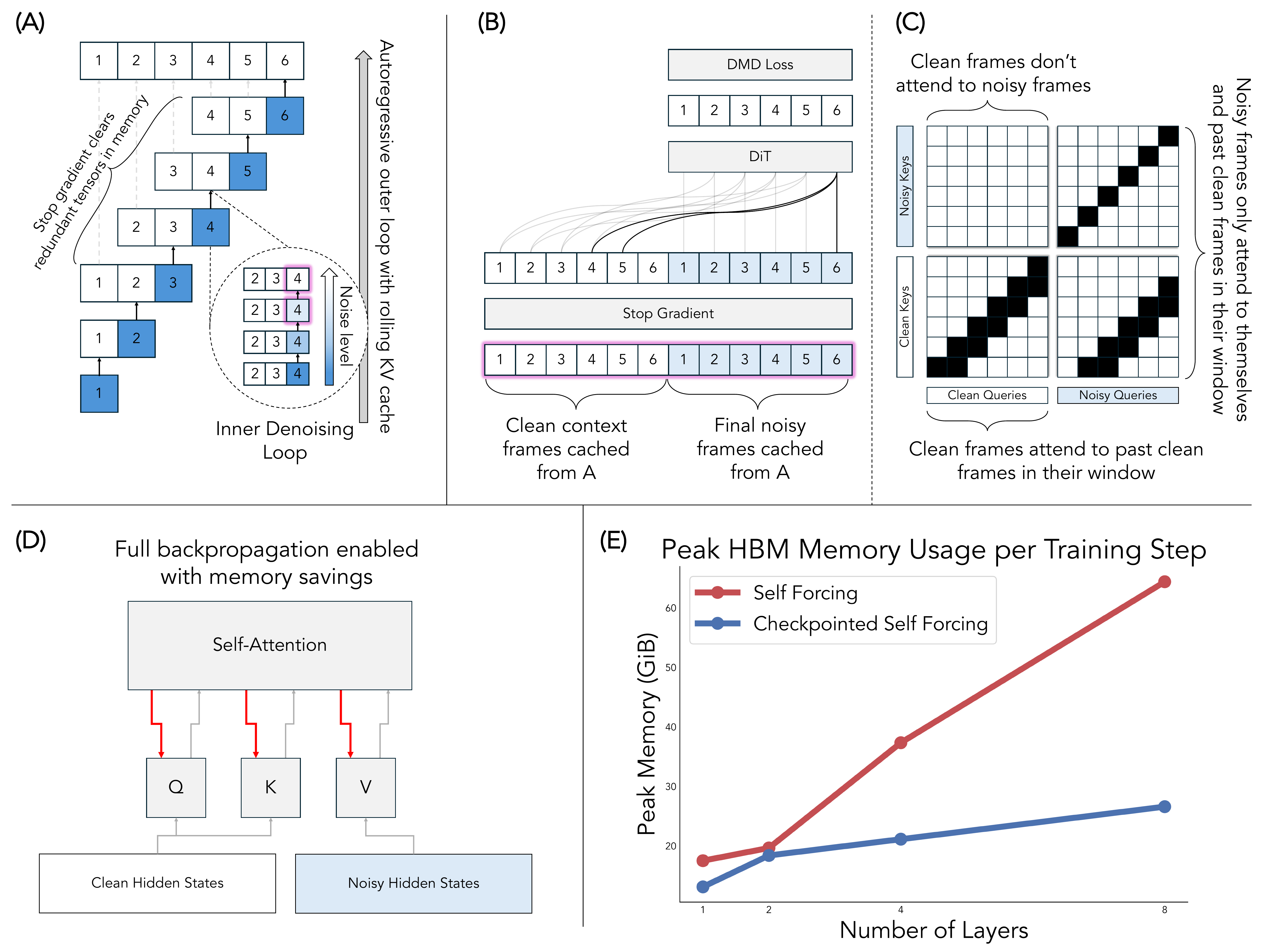}
    \caption{An overview of \oursf. \textbf{(A)} A video is generated with a sliding-window KV cache. The computation graph in Jax contains redundant tensors, making naive backpropagation memory prohibitive. \textbf{(B)} Our recomputation step simulates the last step of denoising for every frame in parallel. Clean context frames and final noisy frames are cached from the previous step. \textbf{(C)} An illustration of the attention mask used in part B (black squares denote attention). This attention mask allows us to simulate the final denoising step of the rollout in parallel. \textbf{(D)} Leveraging these memory savings, we enable backpropagation through KV layers (avoiding the standard stop-gradient of Self Forcing), which we show improves visual generation in~\cref{sec:experiments}. Specifically, we allow gradients to flow in line 29 of~\cref{alg:self_forcing_algorithms}. \textbf{(E)} Peak memory comparison between naive and \oursf across varying depths. Scaled-down networks are used to prevent OOM errors in the naive baseline. }
    \label{fig:self_forcing}
\end{figure*}

\subsection{Stage 4: Self Forcing}
\label{subsec:self-forcing}
We apply Self Forcing~\cite{huang2025selfforcingbridgingtraintest} to improve long generations. The original Self Forcing method (see~\cref{subsec:sf_overview} for an overview) requires the context length of the student model to be equal to that of the teacher. We extend the teacher's context to be longer than the student's, allowing the student to benefit from a more powerful teacher. This requires using a sliding-window context when generating the student's video. However, naively applying Self Forcing when using this sliding window leads to excessive memory usage, as shown in~\cref{fig:self_forcing}. We fix this problem with an efficient implementation termed \textit{\oursf}. Our method first produces an initial video and caches intermediate noisy frames with gradient computation disabled. We then recompute the final video from these intermediate states in an additional step with backpropagation enabled. This process saves redundant memory that would otherwise accumulate when doing backpropagation through sliding-window video generation.

The core issue with naively applying Self Forcing in the sliding-window setting is that each generation step produces a new window of context frames, and backpropagation requires retaining all of these windows in memory simultaneously. For a student context length $L_s$ and a total generation length of $L_t$ steps, the sliding window produces overlapping windows (e.g., frames $1{:}L_s$, then $2{:}L_s{+}1$, then $3{:}L_s{+}2$), each sharing all but one frame with its predecessor. This results in a memory cost of $O(L_t \cdot L_s)$. Our method eliminates this redundancy by decoupling the autoregressive rollout from backpropagation.

Specifically, in \oursf, we first perform the autoregressive rollout to obtain the sequence of clean estimates $\hat{\mathbf{x}}_0^{1:N}$ and the corresponding noisy transition states $\mathbf{x}_{\sigma}^{1:N}$, stopping all gradient propagation during this phase. We then recompute the generator's outputs in a single parallelized forward pass. This pattern of using recomputation to save memory is analogous to Gradient Checkpointing~\cite{chen2016trainingdeepnetssublinear,griewank2000revolve}, where we manually create a checkpoint for $\hat{\mathbf{x}}_0^{1:N}$ and $\mathbf{x}_{\sigma}^{1:N}$ during the autoregressive rollout. To strictly reproduce the inference-time conditioning where noisy frames attend to clean history, we double the input sequence length by concatenating the clean frames $\hat{\mathbf{x}}_0^{1:N}$ and the noisy frames $\mathbf{x}_{\sigma'}^{1:N}$. We then apply a custom ``Teacher Forcing'' attention mask that enforces causal, sliding-window dependencies: it forces each noisy frame $\mathbf{x}_{\sigma}^i$ to attend exclusively to the preceding clean frames within its context window of size $L_s$, i.e., $\hat{\mathbf{x}}_0^{i-L_s:i-1}$. A pseudocode visualization of this procedure, which highlights the differences between \oursf and naive Self Forcing, is shown in~\cref{alg:self_forcing_algorithms}.

This formulation effectively converts the sequential rolling cache operation into a single parallel operation, reducing the memory footprint to $O(L_t)$. Due to our memory savings, we find that backpropagating through the recomputed KV representations (line 29 of~\cref{alg:self_forcing_algorithms}), though not part of the original Self Forcing algorithm, is now feasible and can improve generations (see~\cref{sec:experiments} for results).

\begin{figure}[t!]
\begin{algorithm}[H]
  \caption{\oursf}
  \small
  \SetAlgoLined
  \DontPrintSemicolon
  \KwIn{Denoise timesteps $\{t_1, \dots, t_T\}$, teacher context length $L_t$, student context length $L_s$}
  \KwIn{AR diffusion model $G_\theta$ (returns KV embeddings via $G_\theta^\text{KV}$)}

  Initialize model output $\mathbf{X}_0 \gets []$\;
  \added{Initialize noisy inputs $\mathbf{X}_s \gets []$}\;
  Initialize KV cache $\mathbf{KV} \gets []$\;
  Sample $s \sim \text{Uniform}(1, 2, \dots, T)$\;
  \For{$i \gets 1$ \KwTo $L_t$}{
    Initialize $x^i_{t_T} \sim \mathcal{N}(0, I)$\;
    \For{$j \gets T$ \KwTo $s$}{
      \If{$j = s$}{
        \added{$\mathbf{X}_s.\text{append}(x^i_{t_j})$}\;
        Set $\hat{x}^i_0 \gets G_\theta(x^i_{t_j}; t_j, \mathbf{KV})$\;
        $\mathbf{X}_0.\text{append}(\hat{x}^i_0)$\;
        \removed{Cache $\mathbf{kv}^i \gets G_\theta^\text{KV}(\texttt{stop\_grad}(\hat{x}^i_0); 0, \mathbf{KV})$}\;
        \added{Cache $\mathbf{kv}^i \gets G_\theta^\text{KV}(\hat{x}^i_0; 0, \mathbf{KV})$}\;
        $\mathbf{KV}.\text{append}(\mathbf{kv}^i)$\;
        \lIf{$|\mathbf{KV}| > L_s$}{$\mathbf{KV}.\text{pop}(0)$}
      }
      \Else{
        \removed{Set $\hat{x}^i_0 \gets \texttt{stop\_grad}(G_\theta(x^i_{t_j}; t_j, \mathbf{KV}))$}\;
        \added{Set $\hat{x}^i_0 \gets G_\theta(x^i_{t_j}; t_j, \mathbf{KV})$}\;
        Sample $\epsilon \sim \mathcal{N}(0, I)$\;
        Set $x^i_{t_{j-1}} \gets \texttt{euler\_step}(\hat{x}^i_0, \epsilon, t_{j-1})$\;
      }
    }
  }
  \added{$\mathbf{X}_s \gets \texttt{stop\_grad}(\mathbf{X}_s)$}\;
  \added{$\mathbf{X}_0 \gets \texttt{stop\_grad}(\mathbf{X}_0)$}\;
  \added{$M \gets \texttt{TeacherForcingMask}(L_s, L_t)$}\;
  \added{$\mathbf{X}_{\text{in}} \gets [\mathbf{X}_0, \mathbf{X}_s]$}\;
  \added{$\hat{\mathbf{X}}_0 \gets G_\theta(\mathbf{X}_{\text{in}}; [0, t_s], \text{mask}=M)$}\;
  Update $\theta$ via distribution matching loss\;
\label{alg:self_forcing_algorithms}
\end{algorithm}
\caption{Diff-style pseudocode comparing a single generator train step of naively applying sliding-window self forcing to our efficient implementation, \oursf. Green ``+" lines denote lines from our algorithm that have been added while red ``-" denote lines that have been removed. Please also see~\cref{subsec:teacher_forcing_mask} for mask pseudocode and~\cref{fig:self_forcing} for a visualization.}
\end{figure}

\begin{figure*}[h!]
    \centering
    \includegraphics[width=\linewidth]{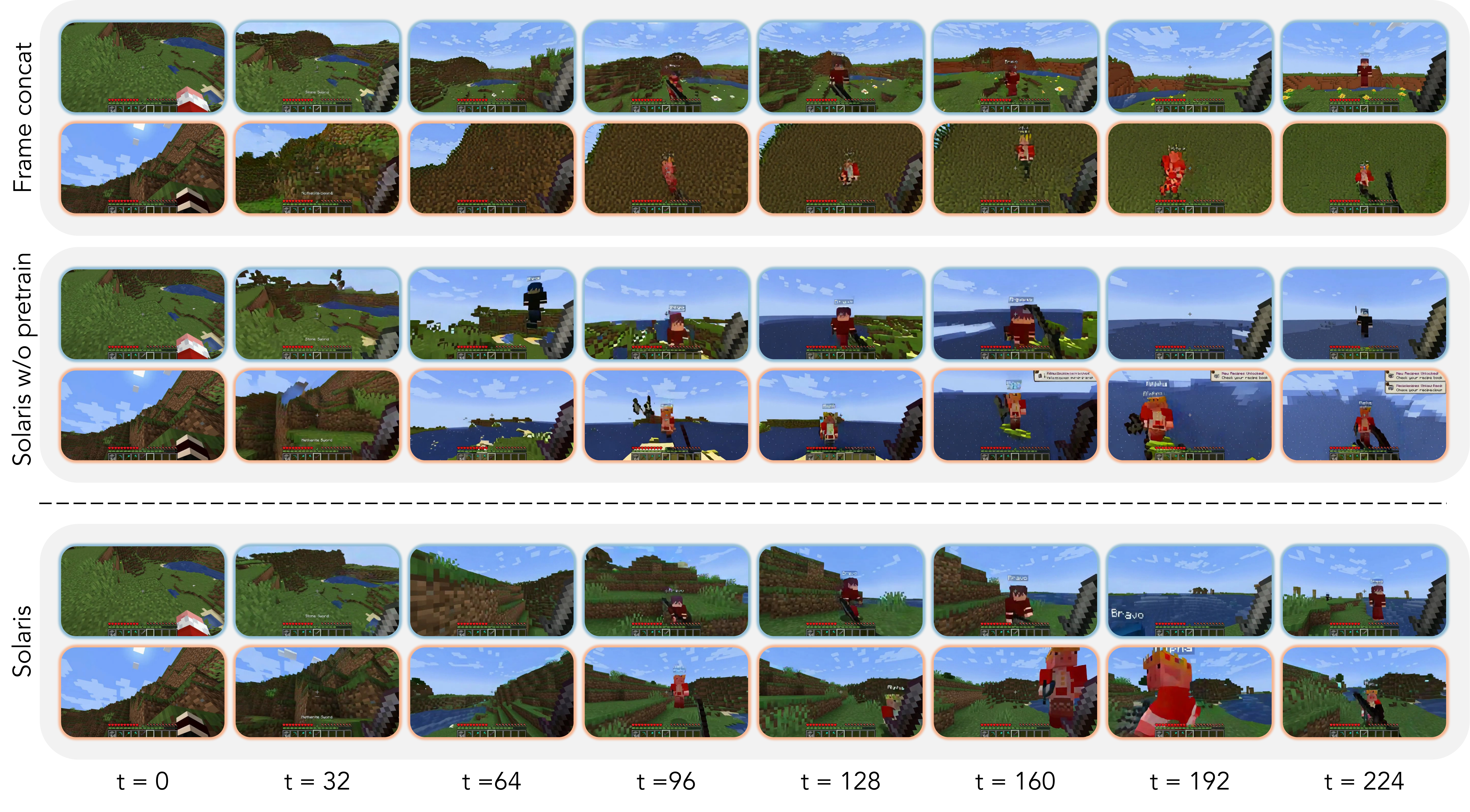}
    \caption{\textbf{Qualitative comparison.} Our model \solaris is able to produce stable and coherent frame generations for long-horizon, as shown here with 224 frames. Unlike the baselines, it maintains realistic fighting gameplay and displays complex terrain that maintains realistic texture. In contrast, the frame concatenation baseline shows severe degradation for the second player and flattened texture for the first player. ``Solaris w/o pretrain" exhibits unnatural behavior such as duplicating player bodies, showing incorrect pop-up notifications, and degenerating to an unrealistic underwater setting.}
    \label{fig:qualitative}
\end{figure*}

\section{Evaluation Benchmark}
\label{sec:eval}

We develop a set of held-out (i.e., completely unseen) episode types that evaluate models on Movement, Grounding, Memory, Building, and Consistency capabilities. We compute FID~\cite{heusel2017gans} to measure overall visual quality and introduce a ``VLM as a judge'' metric for measuring semantic adherence to the task: we ask the VLM judge a verifiable question about a generated video, which varies by task. If its answer matches the known expected answer (e.g. the VLM indeed judges there is a player moving to the left), we count the generated video as accurate. Please see~\cref{sec:appendix_eval} for further details, including text prompts and which frames are sent to the VLM judge. Below, we describe the five evaluation tasks on which we benchmark our models and how the VLM judge is used in each of them.

\textbf{Movement.} We test the model's ability to render visually consistent agent translation (WASD movement) and camera rotation (mouse commands) from both agents' views. In an episode, one bot is moving, and the other one is observing. We use the VLM to judge the position in the observer view.

\textbf{Grounding.} We assess the model's ability to remember the position of the player in the world through the observation of the other player. To this end, we design an episode in the ``Superflat'' world in which two agents are near each other and facing each other. One player turns away and no longer sees the other agent, pauses, and then turns back to its original orientation. Since the turning agent is constantly observed by the stationary player, it should know its position in the world in relation to the other player. The VLM evaluates whether the turning agent sees the other player in its view after it has turned back, and whether the turning agent does not see the other player while it has turned away.

\textbf{Memory.} We assess the model's ability to remember the environment and objects across time. To this end, we design an episode in the ``Superflat'' world in which two agents are near and facing each other. They both turn away from each other, pause, and then return to their original orientations. We ask the VLM whether both agents see the other player in their view after they have turned back, and whether both agents do not see the other player while they have both turned away.

\textbf{Building.} We test the model's ability to reflect environmental changes caused by agents' actions. Specifically, we design a simple block-building episode in the ``Superflat'' world. We begin with two agents near each other. One bot, the builder, builds a simple pre-defined shape (square or strip). The other bot, the observer, watches as the building continues. The players look at the structure and the VLM evaluates whether the observer sees a built structure.

\textbf{Consistency.} We wish to test how well the model understands the correlation between co-visible regions in the two agents' field-of-view. In a real 3D environment, if two agents located near each other turn to look at a previously unseen region, they should see the same region. Therefore, we design an episode in the ``Normal'' world in which two agents are near and facing each other. They then simultaneously turn 90 degrees to either the same side or opposite sides. The VLM judges whether their views are similar (for same-side turning) or different (for opposite-side turning).

\begin{figure*}[h!]
    \centering
    \includegraphics[width=\linewidth]{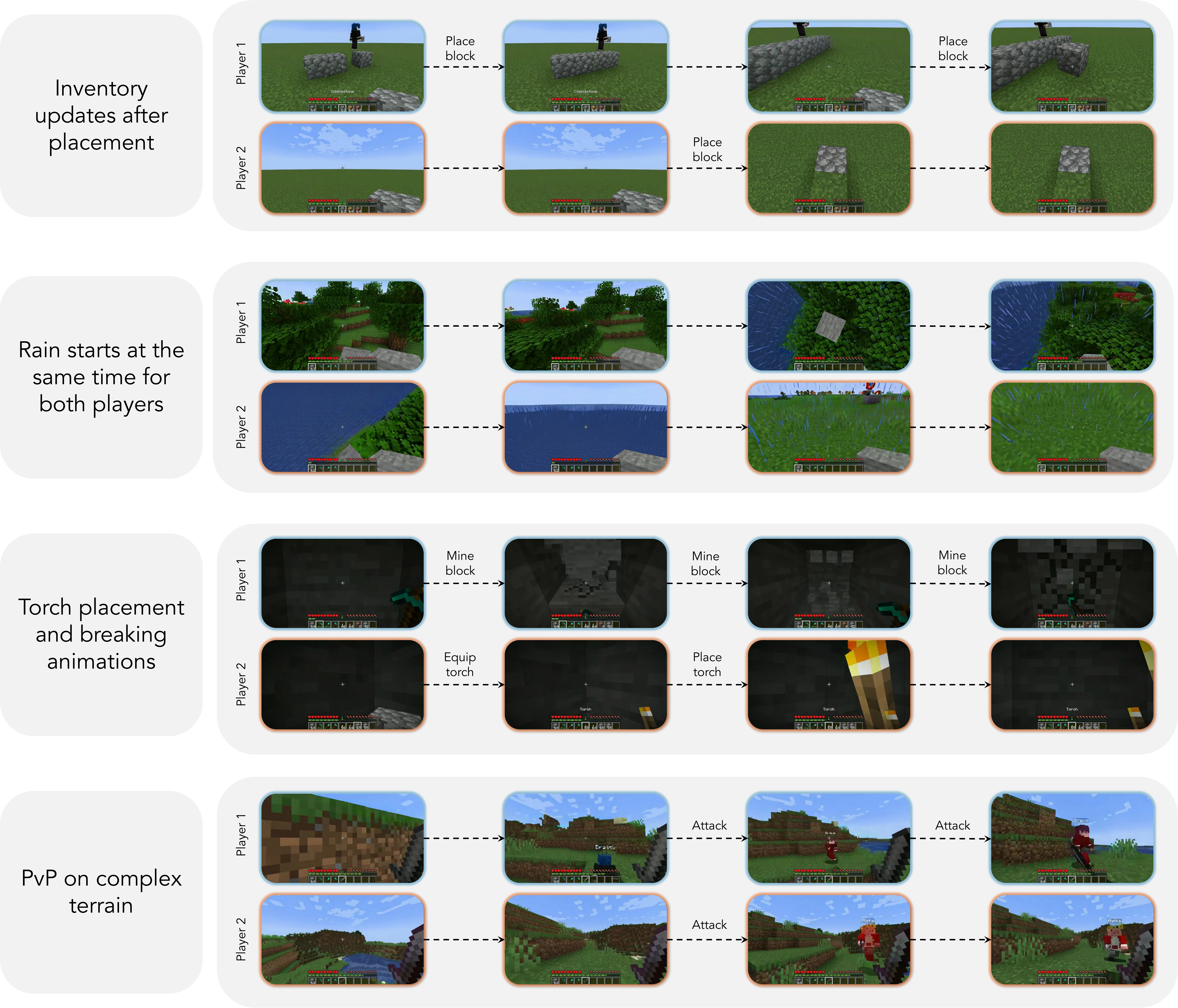}
    \caption{Qualitative examples of learned capabilities. We present generated videos demonstrating the model's ability to simulate complex game dynamics. The rows illustrate: (1) fine-grained state tracking, specifically inventory counter synchronization following block placement; (2) global environmental consistency, shown by the simultaneous onset of rain; (3) inventory active item synchronization, torch placement, and generating accurate mining animations; and (4) coherent PVP on complex terrain.}
    \label{fig:capabilities}
\end{figure*}

\section{Experiments}
\label{sec:experiments}
\begin{table}[t]
\centering

\resizebox{\textwidth}{!}{%
\begin{tabular}{lcccccccccc}
\toprule
 & \multicolumn{2}{c}{\textbf{Movement}} 
 & \multicolumn{2}{c}{\textbf{Grounding}} 
 & \multicolumn{2}{c}{\textbf{Memory}} 
 & \multicolumn{2}{c}{\textbf{Building}} 
 & \multicolumn{2}{c}{\textbf{Consistency}} \\
\cmidrule(lr){2-3} 
\cmidrule(lr){4-5} 
\cmidrule(lr){6-7} 
\cmidrule(lr){8-9} 
\cmidrule(lr){10-11}
\textbf{Method} 
& \textbf{VLM} $\uparrow$ & \textbf{FID} $\downarrow$ 
& \textbf{VLM} $\uparrow$ & \textbf{FID} $\downarrow$ 
& \textbf{VLM} $\uparrow$ & \textbf{FID} $\downarrow$ 
& \textbf{VLM} $\uparrow$ & \textbf{FID} $\downarrow$ 
& \textbf{VLM} $\uparrow$ & \textbf{FID} $\downarrow$ \\
\midrule
Frame concat & \textbf{77.1 $\pm$ 2.9} & 68.9 & 53.1 $\pm$ 0.0 & 66.6 & \textbf{37.5 $\pm$ 2.6} & 74.4 & 0.0 $\pm$ 0.0 & 103.2 & 49.5 $\pm$ 5.3 & 129.4 \\
\solaris w/o pretrain  & 69.3 $\pm$ 0.7 & 42.5 & 29.2 $\pm$ 1.5 & 49.9 & 18.8 $\pm$ 0.0 & 67.8 & 0.0 $\pm$ 0.0 & 86.6 & 49.5 $\pm$ 3.2 & 121.4 \\
\solaris & 68.2 $\pm$ 2.7 & \textbf{38.5} & \textbf{62.5 $\pm$ 0.0} & \textbf{38.0} & \textbf{37.5 $\pm$ 2.6} & \textbf{55.1} & \textbf{20.8 $\pm$ 1.5} & \textbf{83.6} & \textbf{71.4 $\pm$ 7.1} & \textbf{99.4} \\
\bottomrule
\end{tabular}%
} %

\caption{Quantitative comparison across tasks. We compare our method against concatenating player observations along the channel dimension following Enigma Multiverse~\cite{enigma2025multiverse} and training from scratch without single-player pretraining. ``VLM'' shows the VLM accuracy $\pm$ 1 (estimated) standard deviation. The standard deviation is estimated by repeating the VLM eval 3 times.}
\label{tab:arch_ablation_results}
\end{table}

\subsection{Qualitative Results}
Here we discuss qualitative results from our flagship, \solaris, model. As shown in~\cref{fig:teaser}, our model is capable of simulating complex aspects of Minecraft gameplay, including building, mining, fighting, and multiplayer viewpoint modeling. The action sequences for each of these scenarios come from bot gameplay not present in the training dataset. We also recorded a third-person perspective of the bots, shown in the left and right columns. As can be seen in the end frames from our model, its generations align with the ground truth game state. Examples of advanced model capabilities---including inventory tracking, simulating weather, placing torches, generating animations, and simulating PvP---are presented in~\cref{fig:capabilities}. ~\cref{fig:qualitative} presents qualitative results from our architecture comparison discussed in~\cref{subsec:architecture_experiments}. While baseline methods degrade over time, our model maintains visual fidelity throughout a long-horizon.

\subsection{Architecture Experiments}
\label{subsec:architecture_experiments}

We compare our architecture implementation to the frame concatenation method of Multiverse ~\cite{enigma2025multiverse}, the only existing multiplayer world model prior to this work. We also test the necessity of single-player pretraining by comparing the performance of our method to the variant trained without the single-player model initialization. Our method produces superior visual results both qualitatively, as shown in~\cref{fig:qualitative}, and quantitatively across all evaluation categories, as shown in~\cref{tab:arch_ablation_results}. All architecture variants are strong at action following in motion-based trajectories, and get a high VLM score on the evaluation in the corresponding category, as shown in~\cref{tab:arch_ablation_results}. Our method shows superior performance on difficult scenarios involving building, scene consistency, and player grounding, reflected by the higher VLM scores in those categories. Although the frame concatenation method outperforms in our Movement evaluation, we find qualitatively that it exhibits action hallucinations in the presence of no-op actions. 

\subsection{Self-Forcing Ablations}
\label{subsec:sf_ablations}

We ablate each component of our Self Forcing pipeline. We study the two main stages of CausVid, ODE regression initialization and few-step distillation, and find that straightforward causal finetuning is sufficient instead. Although the original Self Forcing paper assumes the generator to be a few-step model at the start of training, we find that the few-step ability can be learned simultaneously with stable autoregressive generation in Self Forcing. This allows use simple finetuning instead of CausVid.

Next, we study the ability to backpropagate to the KV representations of the self attention layer, which is feasible with the memory savings of our method. Allowing KV backpropagation achieves better visual results than all other variants based on FID, as shown in~\cref{tab:sf_ablation_results}. We do observe that this causes decreased performance in action following for some categories. However, our method remains competitive across all categories and excels in the challenging Building and Consistency VLM tasks.

\begin{table*}[h!]
\centering

\resizebox{\textwidth}{!}{%
\begin{tabular}{lcccccccccccc}
\toprule
\multicolumn{3}{c}{\textbf{Components}} & 
\multicolumn{2}{c}{\textbf{Movement}} & 
\multicolumn{2}{c}{\textbf{Grounding}} & 
\multicolumn{2}{c}{\textbf{Memory}} & 
\multicolumn{2}{c}{\textbf{Building}} & 
\multicolumn{2}{c}{\textbf{Consistency}} \\
\cmidrule(r){1-3}
\cmidrule(lr){4-5}
\cmidrule(lr){6-7}
\cmidrule(lr){8-9}
\cmidrule(lr){10-11}
\cmidrule(l){12-13}
\textbf{Init.} & \textbf{Pre-DMD} & \textbf{KV-BP} &
\textbf{VLM} $\uparrow$ & \textbf{FID} $\downarrow$ &
\textbf{VLM} $\uparrow$ & \textbf{FID} $\downarrow$ &
\textbf{VLM} $\uparrow$ & \textbf{FID} $\downarrow$ &
\textbf{VLM} $\uparrow$ & \textbf{FID} $\downarrow$ &
\textbf{VLM} $\uparrow$ & \textbf{FID} $\downarrow$ \\
\midrule
ODE Reg & \texttimes & \checkmark & 23.4 $\pm$ 1.5 & 65.3 & 3.1 $\pm$ 0.0 & 56.6 & 0.0 $\pm$ 0.0 & 99.5 & 3.1 $\pm$ 0.0 & 95.7 & 49.0 $\pm$ 4.2 & 142.3 \\
Causal FT & \checkmark & \checkmark & 21.4 $\pm$ 2.7 & 49.1 & 3.1 $\pm$ 0.0 & 40.4 & 0.0 $\pm$ 0.0 & 55.7 & 8.3 $\pm$ 1.5 & 90.5 & 55.2 $\pm$ 3.7 & 160.1 \\
Causal FT & \texttimes & \texttimes & \textbf{78.6 $\pm$ 0.7} & 60.3 & \textbf{72.9 $\pm$ 1.5} & 55.2 & \textbf{49.0 $\pm$ 2.9} & 63.8 & 15.6 $\pm$ 0.0 & 87.4 & 70.8 $\pm$ 6.6 & 105.1 \\
\rowcolor{gray!15} Causal FT & \texttimes & \checkmark & 68.2 $\pm$ 2.7 & \textbf{38.5} & 62.5 $\pm$ 0.0 & \textbf{38.0} & 37.5 $\pm$ 2.6 & \textbf{55.1} & \textbf{20.8 $\pm$ 1.5} & \textbf{83.6} & \textbf{71.4 $\pm$ 7.1} & \textbf{99.4} \\
\bottomrule
\end{tabular}%
}

\caption{Ablations on Self Forcing training variants. We study the initialization of the causal model, finding simple causal finetuning with Diffusion Forcing~\cite{chen2024diffusionforcingnexttokenprediction} to suffice. We also find that doing few-step distillation before Self-Forcing is ineffective. Finally, we find that enabling KV cache backpropagation improves visual quality. ``VLM'' shows the VLM accuracy $\pm$ 1 (estimated) standard deviation. The standard deviation is estimated by repeating the VLM eval 3 times.}
\label{tab:sf_ablation_results}
\end{table*}

\section{Conclusion}

We present \solaris, a multiplayer video world model capable of generating consistent multi-view observations from coordinated multi-agent interactions. By developing a scalable multiplayer data collection system, a staged training pipeline transitioning from single-player to multiplayer modeling, and \oursf for memory-efficient long-horizon training, we demonstrated that coherent multi-agent world simulation is achievable.

While this paper leveraged \solarisengine to collect two-player Minecraft data, the platform's potential extends well beyond this setting. It naturally supports more than two concurrent players and can serve a variety of downstream research directions: generating multimodal training data for vision-language models or vision-language-action models, training unified models that jointly perceive and act, developing single-agent and multi-agent policies, studying neurosymbolic approaches where agents reason through code, and constructing benchmarks for 3D understanding, planning, and memory.

Several limitations of our current work also point to promising avenues for future research. First, our training data is entirely synthetic, which introduces gaps in both the action and visual distributions the model encounters, limiting its ability to generalize. While we have demonstrated the value of single-player pretraining, future work could further investigate how to leverage the more abundant single-player data to close this gap. Second, our model lacks persistent memory: when players leave each other's field of view, the model loses track of their shared context, and their trajectories begin to diverge. Unlike a true game engine, the world has no underlying persistent state, as it is specified only through two initial frames, with no mechanism to control or maintain the broader environment. Our work is a valuable starting point for addressing these future research challenges. 

\section{Acknowledgments}
Srivats Poddar completed this work while studying at NYU. Suppakit Waiwitlikhit and Timothy Meehan contributed to the project during their time as visiting students at NYU. We thank Nanye Ma for his help with our Jax codebase. We are grateful for Jihan Yang, Sihyun Yu, Shusheng Yang, and Yucen Lily Li for their advice on the draft and Charles Herrmann for helpful discussions. Egor Gikalo made very useful improvements to the \solarisengine codebase. This work was primarily supported by the Google TPU Research Cloud (TRC) program and the Google Cloud Research Credits program (GCP19980904). S.X. acknowledges support from the MSIT IITP grant (RS-2024-00457882) and the NSF award IIS-2443404. O.M. is supported by the NSF Graduate Research Fellowship Program. 
\newpage

\addcontentsline{toc}{section}{References}
\bibliography{main}
\bibliographystyle{plain}

\clearpage
\appendix
\section{\solaris Multiplayer Framework}
\label{sec:data_engine_details}

\subsection{Mineflayer Modifications}
To facilitate a wide range of Minecraft action recording, we modify the Mineflayer API to expose access to the most recently applied camera action in its physics module and extend its event system to send events on one-off semantic actions such as attacking, using, placing, and hotbar changes.

In addition, we introduce the following Mineflayer code to improve the quality of the collected data: 
\begin{itemize}[noitemsep, topsep=0pt]
  \item Make the bot correctly look at the face of the block when placing a new block.
  \item Add camera smoothing to all non-Pathfinder look commands.
\end{itemize}

\subsection{GPU Data Collection}

We initially experimented with \texttt{libOSMesa} for CPU-based headless rendering for the Minecraft clients (camera bots), however, videos in the open, \textit{Normal} world often appeared laggy, having up to four repeating frames at a 20 fps recording in graphics-intensive scenes, such as in a forest. Switching to GPU-based rendering produced consistently smooth videos without frame repetition. Additionally, switching to the GPU enabled the use of NVENC, Nvidia's hardware video encoder, which further reduced CPU workload as it had to encode screen captures via \texttt{ffmpeg}.

\section{Multiplayer Training Dataset}
\label{sec:appendix_train_dset}
\subsection{Episode Types}
\label{subsec:appendix_episode_types}

Our dataset contains 14 distinct episode types that cover the 4 broader Minecraft game mechanics. We describe the episode types in~\cref{tab:episode_types_details}.

\begin{table}[h]
\centering
\small
\newcolumntype{Y}{>{\raggedright\arraybackslash}X}

\caption{Episode types in our training dataset.}
\label{tab:episode_types_details}

\begin{tabularx}{\linewidth}{llY}
\toprule
\textbf{Episode Name} & \textbf{Category} & \textbf{Description} \\
\midrule
buildStructure & Building & Each bot builds a wall or tower in front of the other, or bots build a platform together at midpoint \\
buildTower & Building & Both bots build a tall 1-block tower by jumping and placing blocks underneath themselves. \\
buildHouse & Building & Bots build a house together. \\
towerBridge & Building & Bots build a 1-block tower by jumping, then build a bridge connecting the two towers, and meet. \\
pve & Combat & Bots defend two random positions on the ground, facing each other against spawning mobs. \\
pvp & Combat & Bots equip swords and fight each other. \\
collector & Mining & One bot digs underground to search and mine ores within line-of-sight, while the other bot follows and places torches. \\
mine & Mining & Bots dig 1 block underground and mine their way towards each other. \\
placeAndMine & Mining & Bots stand facing each other. One bot places blocks, and the other one destroys them. \\
chase & Movement & One bot runs away in a zig-zag pattern and the other bot pursues it. \\
orbit & Movement & Bots move in a circular trajectory around a shared center, periodically looking at each other. \\
straightLineWalk & Movement & One bot runs towards and past the other bot, then spins to look at it. \\
walkLook & Movement & One or both bots move in a random direction with just WASD actions in front of each other. \\
walkLookAway & Movement & One bot moves in a random direction, looks away, looks back at the other bot. The other observes. \\
\bottomrule
\end{tabularx}
\end{table}

\subsection{Episode filtering}

\begin{figure}[h]
    \centering
    \includegraphics[width=\linewidth]{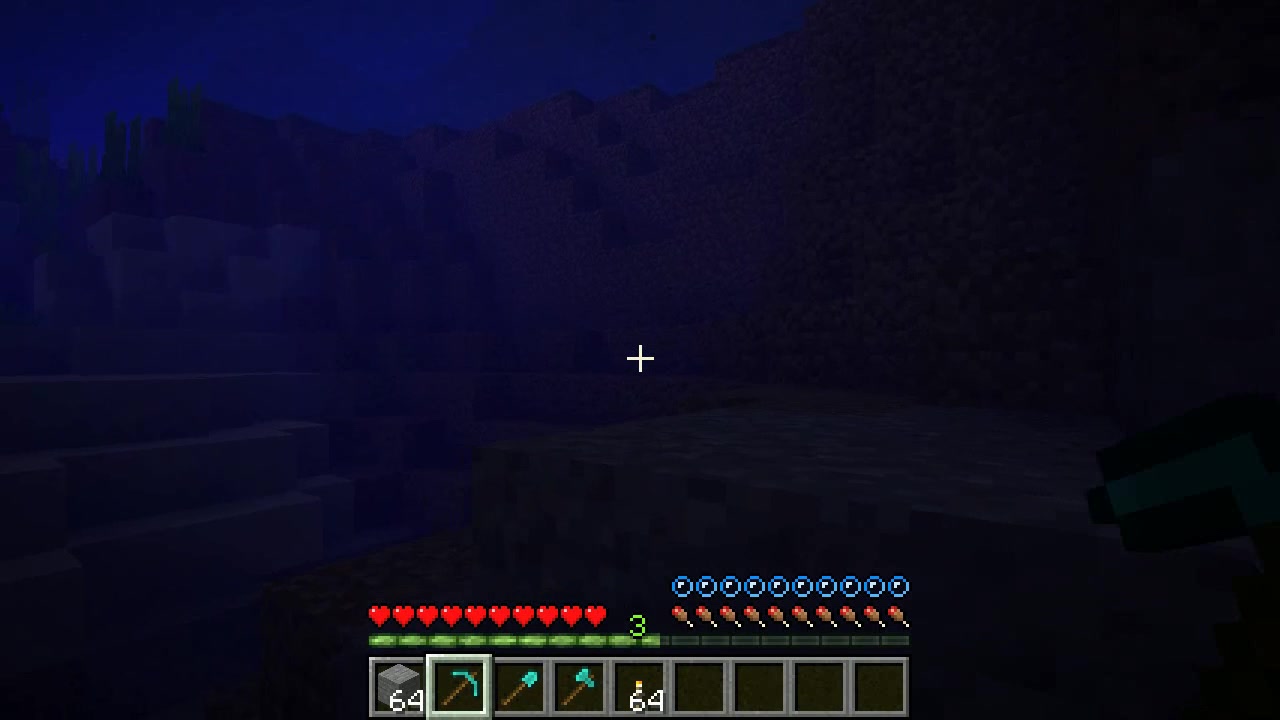}
    \caption{An episode where a bot has been teleported underwater.}
    \label{fig:underwater_episode}
\end{figure}

\begin{figure}[h]
    \centering
    \includegraphics[width=\linewidth]{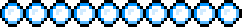}
    \caption{The ``bubbles'' HUD template PNG used to filter underwater episodes.}
    \label{fig:bubbles_gui}
\end{figure}

Due to the teleportation logic we use before every episode to diversify the terrain, the bots might end up underwater in some of the episodes, as shown in~\cref{fig:underwater_episode}. The underwater mechanics break the logic of most of our episodes, so we choose not to include those. Out of 6000 episodes collected in the ``Normal'' world, 340 are underwater. Leveraging the fact that Minecraft shows oxygen bubbles in the GUI, see~\cref{fig:bubbles_gui}, when the character is underwater, we filter out the water episodes using a linear classifier that we run on the episode frames, detecting the bubble icons. The classifier achieves a 100\% accuracy.

\subsection{Action Space}

\label{subsec:appendix_action_space}

The gold standard of actions for a Minecraft dataset for world modeling is the VPT ~\cite{baker2022videopretrainingvptlearning} dataset, which contains the full range of Minecraft game actions. \solarisengine and hence our multiplayer dataset come close to it and support everything except for inventory opening/closing and raw mouse movement recording (only relative pitch/yaw character rotations are recorded). These two limitations don't allow \solarisengine to capture GUI activities such as inventory manipulation and crafting. Another difference between VPT and our multiplayer dataset is that we record and store actions in the form of semantic actions (either continuous or discrete) instead of raw keyboard and mouse events, as the VPT dataset does. The full list of actions recorded by \solarisengine and available in our multiplayer training dataset is presented in ~\cref{tab:action_space}.

\begin{table}[h]
\centering
\small
\newcolumntype{Y}{>{\raggedright\arraybackslash}X}
\caption{Action space of our multiplayer dataset. ``Sustained'' denotes action is recorded for as long as it lasts. ``Once'' denotes action is recorded for one tick only, usually applicable to instantaneous events.}
\label{tab:action_space}

\begin{tabularx}{\linewidth}{llY}
\toprule
\textbf{Action key} & \textbf{Type} & \textbf{Description} \\
\midrule
forward       & bool/sustained & Player moving forward (W). \\
back          & bool/sustained & Player moving backward (S). \\
left          & bool/sustained & Player strafing left (A). \\
right         & bool/sustained & Player strafing right (D). \\
jump          & bool/sustained & Player jumping. \\
sprint        & bool/sustained & Player sprinting. \\
sneak         & bool/sustained & Player sneaking. \\
camera        & vec2f/sustained & Change in player camera orientation (yaw, pitch). \\
attack        & bool/once & Player attacks. \\
use           & bool/once & Player uses / interacts with the environment. \\
mount         & bool/once & Player mounts an entity/vehicle. \\
dismount      & bool/once & Player dismounts. \\
place\_block  & bool/once & Player places a block using the currently selected item. \\
place\_entity & bool/once & Player places an entity item. \\
mine          & bool/sustained & Player mining a block. \\
hotbar.1      & bool/once & Player selects hotbar slot 1. \\
hotbar.2      & bool/once & Player selects hotbar slot 2. \\
hotbar.3      & bool/once & Player selects hotbar slot 3. \\
hotbar.4      & bool/once & Player selects hotbar slot 4. \\
hotbar.5      & bool/once & Player selects hotbar slot 5. \\
hotbar.6      & bool/once & Player selects hotbar slot 6. \\
hotbar.7      & bool/once & Player selects hotbar slot 7. \\
hotbar.8      & bool/once & Player selects hotbar slot 8. \\
hotbar.9      & bool/once & Player selects hotbar slot 9. \\
\bottomrule
\end{tabularx}
\end{table}

\subsection{Mouse Action Distribution}

\begin{figure}[h]
    \centering
    \includegraphics[width=\linewidth]{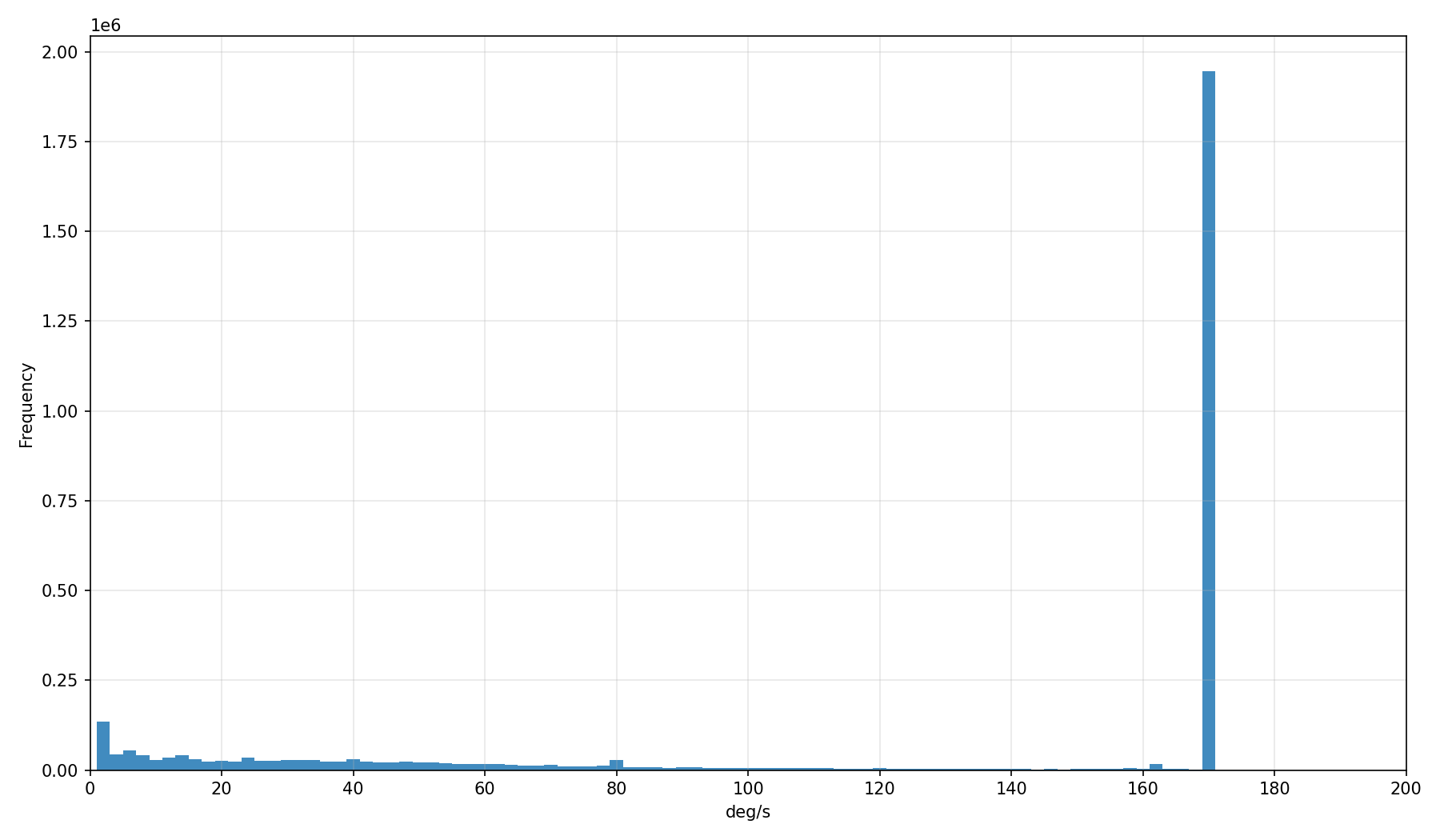}
    \caption{A histogram showing the mouse action magnitude distribution in our training dataset.}
    \label{fig:mouse_act_histogram}
\end{figure}

Most of our episode types use the Pathfinder Mineflayer plugin for intelligent navigation of our bots. Pathfinder operates at the maximum camera speed of 178 degrees per second. Although we use manual camera movements (not controlled by Pathfinder) in our dataset, which provides camera action at random, lower value ranges, the distribution of camera action of our dataset is heavily skewed towards the fast camera moves originating in Pathfinder, as shown in~\cref{fig:mouse_act_histogram}.

\section{Model Training}

We provide our training hyperparameters in~\cref{tab:hyperparameters}. We train our Bidirectional Single-player model on v5p-128 and all other models on v5p-64 TPUs.

\begin{table}[h]
    \centering
    \captionsetup{type=table}
    \captionof{table}{Training hyperparameters across all training stages.}
    \label{tab:training_hparams}
    \resizebox{\linewidth}{!}{%
        \begin{tabular}{lcccccc}
        \toprule
        \textbf{Setting} & \textbf{Learning Rate} & \textbf{Batch Size} & \textbf{Num Steps} & \textbf{Adam $\beta_1$} & \textbf{Adam $\beta_2$} & \textbf{Weight Decay} \\
        \midrule
        Bidirectional Single-player  & $1\mathrm{e}{-4}$ & 64 & 120K  & 0.9 & 0.95 & 0 \\
        Bidirectional Multiplayer    & $1\mathrm{e}{-4}$ & 32 & 120K  & 0.9 & 0.95 & 0 \\
        Causal Multiplayer           & $1\mathrm{e}{-4}$ & 32 & 60K   & 0.9 & 0.95 & 0 \\
        Self Forcing (generator)     & $3\mathrm{e}{-6}$ & 32 & 240   & 0.9 & 0.95 & 0 \\
        Self Forcing (critic)        & $3\mathrm{e}{-7}$ & 32 & 1200  & 0.9 & 0.95 & 0 \\
        \bottomrule
        \end{tabular}%
    }
    \label{tab:hyperparameters}
\end{table}

\section{Evaluation Details}
\label{sec:appendix_eval}
\begin{figure*}[h!]
    \centering
    \includegraphics[width=\linewidth]{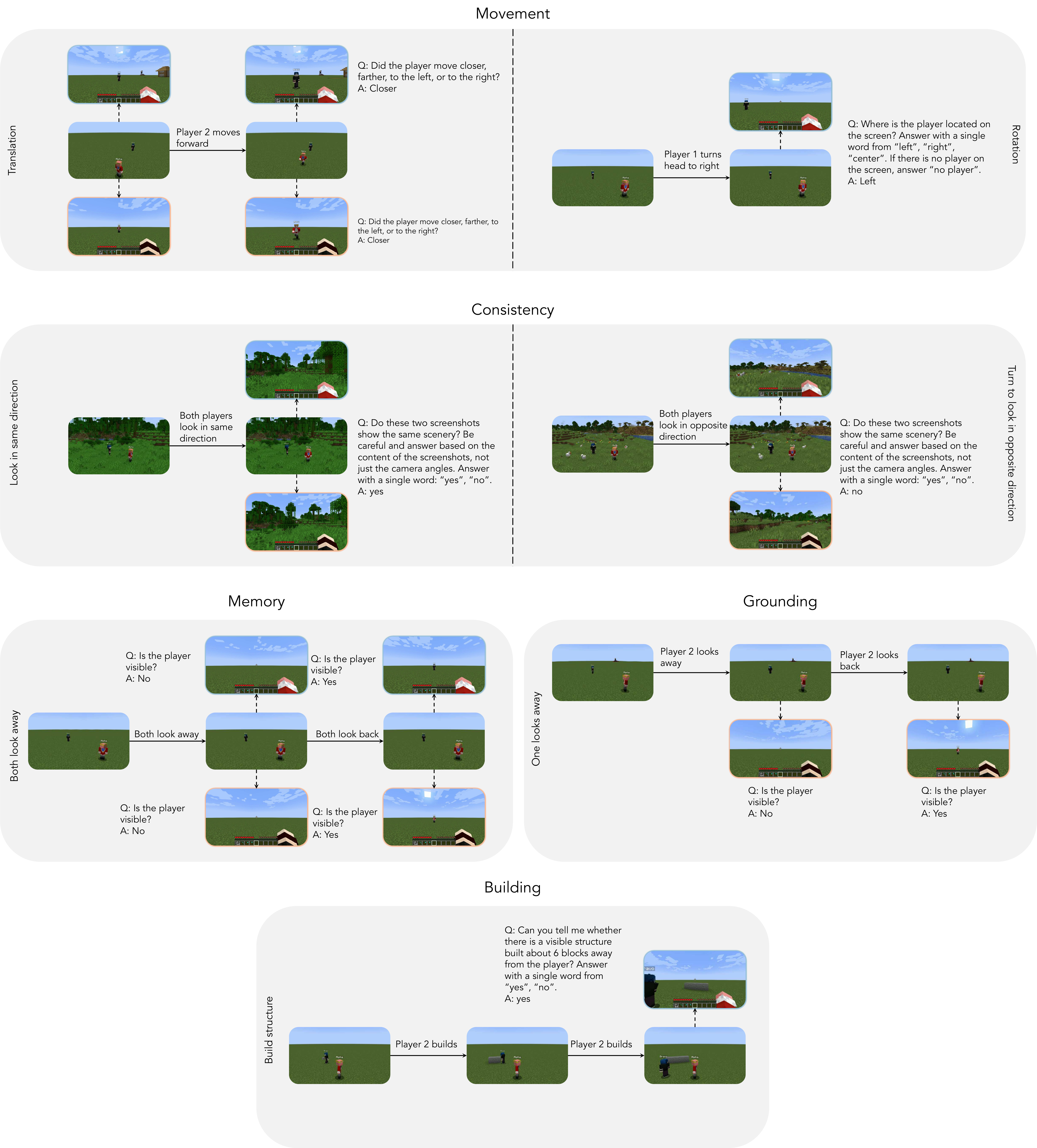}
    \caption{An illustration of the frames and prompts provided to the VLM for evaluation. Each scenario is a ground truth video for illustration. Above the third-person visualization, we show the frames that are sent to the VLM along with the corresponding question and correct answer. Prompts are abbreviated; please see~\cref{tab:eval_vlm_queries} for the full text.}
    \label{fig:queries}
\end{figure*}

In this section, we present the implementation details of our VLM-as-a-judge evaluation method, providing the exact VLM prompts and its accuracy on ground truth videos.  The VLM accuracy is calculated based on whether the VLM gives the expected answer for all queries across all perspectives for the same episode. E.g., for ``Memory'', where both players turn away, an episode is judged as accurate if and only if the VLM's answer matches the expected answer at both query points and for both perspectives ($2 \times 2$ total queries). By contrast, for ``Consistency'', there is one query per episode, but it includes both players' perspectives. Please see~\cref{tab:eval_vlm_queries} for the exact prompt and accuracy on ground-truth episodes (“Episode GT Acc.”), and~\cref{fig:queries} for a visualization.

  \begin{table}[h!]
  \centering                                                                                                                                                                                                         
  \small                                                          
  \setlength{\tabcolsep}{3pt}
  \setlength{\extrarowheight}{3pt}
  \begin{adjustbox}{max width=\linewidth,center}

\begin{tblr}{
  colspec = {|c|m{2.2cm}|m{2.4cm}|m{0.32\linewidth}|m{1.6cm}|m{1.8cm}|},
  row{1} = {font=\bfseries}, %
  hlines, vlines, vspan = even %
}
Task & Ep. Type & {Query Point,\\Perspective} & Prompt & {Exp.\\Answer(s)} & {GT Ep. Acc.\\$\pm$ 1 SD (\%)} \\

\SetCell[r=2]{c} Movement 
  & Translation & {Movement Ends,\\Each Player} & Here are Minecraft screenshots showing another player on the screen. Between the first frame and the second frame, did the player being shown move closer, farther, to the left, or to the right on-screen? Answer with a single word from ``closer'', ``farther'', ``left'', ``right'', or ``no motion''. & {closer,\\farther,\\left,\\right} & 100.00 $\pm$ 0 \\
  & Rotation & {Turning Ends,\\Acting Player} & Here is a Minecraft screenshot potentially showing another player on the screen. Where is the player located on the screen? Answer with a single word from ``left'', ``right'', ``center''. If there is no player on the screen, answer ``no player''. & {left,\\right,\\center} & 96.88 $\pm$ 0 \\

\SetCell[r=2]{c} Grounding 
  & \SetCell[r=2]{l}{One Player\\Turns Away} & {Turned away,\\Acting Player} & \SetCell[r=2]{l} Here is a Minecraft screenshot. Is there another player visible on-screen? Answer with a single word: ``yes'', ``no''. & no & \SetCell[r=2]{l} 96.88 $\pm$ 0 \\
  & & {Turned back,\\Acting Player} & & yes & \\

\SetCell[r=2]{c} Memory 
  & \SetCell[r=2]{l}{Both Players\\Turn Away} & {Turned away,\\Each Player} & \SetCell[r=2]{l} Here is a Minecraft screenshot. Is there another player visible on-screen? Answer with a single word: ``yes'', ``no''. & no & \SetCell[r=2]{l} 92.71 $\pm$ 1.47 \\
  & & {Turned back,\\Each Player} & & yes & \\

Building & Building & {Building Ends,\\Observing Player} & Here is a Minecraft screenshot. Can you tell me whether there is a visible structure built about 6 blocks away from the player? Answer with a single word from ``yes'', ``no''. & yes & 98.96 $\pm$ 1.47 \\

\SetCell[r=2]{c} Consistency 
  & {Turn $90^\circ$\\ (Same Side)} & \SetCell[r=2]{l} {Turning Ends,\\ Both Players} & \SetCell[r=2]{l} You will be shown two Minecraft screenshots. Do these two screenshots show the same scenery? Be careful and answer based on the content of the screenshots, not just the camera angles. Answer with a single word: ``yes'', ``no''. & yes & 98.96 $\pm$ 1.47 \\
  & {Turn $90^\circ$\\ (Opposite Sides)} & & & no & 93.75 $\pm$ 4.42 \\
\end{tblr}
  \end{adjustbox}
  \caption{Detailed description for our evaluation episodes. The ``GT Ep. Acc.'' column shows the VLM accuracy for ground-truth episodes (which is expected to be $100\%$). In all tasks except for ``Memory'' and ``Consistency'' (where both players turn), one ``acting'' player performs actions while the
  other ``observing'' player stands completely still. In the Query Perspective column, ``Each Player'' means we ask the VLM the same text prompt twice, each time about a different player's perspective (2 queries). ``Both
  Players'' means we ask the VLM judge one text prompt, but about both player's perspectives (1 query). Standard deviation calculated across 3 trials.}
  \label{tab:eval_vlm_queries}
  \end{table}

\section{Self Forcing}

\subsection{Overview}
\label{subsec:sf_overview}

Here, we give a brief overview of the original implementation of the Self Forcing algorithm. Let $N$ be the number of frames in the video sequence. We employ a few-step denoising strategy with a discretized noise schedule $\{\sigma_0=0, \dots, \sigma_K=1\}$. 
A core component of Self Forcing is the ability to obtain a clean prediction from any intermediate noisy state. In Flow Matching, given a noisy latent $\mathbf{x}_{\sigma}$ and the vector field prediction $v_\theta(\mathbf{x}_{\sigma}, \sigma)$, the estimated clean data $\hat{\mathbf{x}}_0$ can be derived analytically: $\hat{\mathbf{x}}_0 = \mathbf{x}_{\sigma} - \sigma v_\theta(\mathbf{x}_{\sigma}, \sigma)$.

During training, fully unrolling the denoising process for every frame is memory-prohibitive. To address this, Self Forcing employs a \textit{stochastic gradient truncation} strategy. For each training iteration, a random target noise level $\sigma_{stop} \sim \{\sigma_0, \dots, \sigma_K\}$ is sampled. The model generates the video autoregressively, but for each frame, the denoising process is truncated once it reaches $\sigma_{stop}$. The input to the next frame $i$ is conditioned on the clean estimates of the previous frames, derived from this truncated state:
\[ \hat{\mathbf{x}}_0^{1:N} \sim p_{\theta} = \prod_{i=1}^N p_{\theta}(\hat{\mathbf{x}}_0^i \mid \hat{\mathbf{x}}_0^{<i}). \]
The DMD loss is applied to the sequence of estimates $\hat{\mathbf{x}}_0^{1:N}$. Gradients are backpropagated only from this final truncation step, and in the original implementation, a stop-gradient operation is added to the KV cache. This random sampling of $\sigma_{stop}$ ensures that the model receives supervision across all noise levels while maintaining constant memory complexity with respect to the number of diffusion steps.

\subsection{Teacher Forcing Mask Implementation}
\label{subsec:teacher_forcing_mask}

In \oursf, a special teacher forcing mask is used to re-simulate the final step of denoising across all frames in parallel. This mask enforces sliding window causality and that noisy frames can only attend to past clean frames. We present pseudocode in~\cref{alg:teacher_forcing_mask}.

\begin{algorithm}[t]
\DontPrintSemicolon
\caption{Pseudocode of \texttt{TeacherForcingMask} in a NumPy-like style.}
\label{alg:teacher_forcing_mask}
\definecolor{codeblue}{rgb}{0.25,0.5,0.5}
\lstset{
  basicstyle=\fontsize{7.2pt}{7.2pt}\ttfamily\selectfont,
  commentstyle=\fontsize{7.2pt}{7.2pt}\color{codeblue}\ttfamily\selectfont,
  keywordstyle=\fontsize{7.2pt}{7.2pt}\ttfamily\selectfont,
}
\begin{lstlisting}[language=python]
# L_s: student context length (sliding window size)
# L_t: teacher context length (number of frames)
# tokens_per_frame: number of tokens per frame

ctx_len = L_t * tokens_per_frame

# create index grids for queries and keys
q_idx = np.arange(2 * ctx_len)[:, None]
kv_idx = np.arange(2 * ctx_len)[None, :]

# compute frame indices
q_frame = (q_idx // tokens_per_frame) %
kv_frame = (kv_idx // tokens_per_frame) %

# determine if query/key is from noisy or clean frames
q_is_noisy = q_idx >= ctx_len
kv_is_noisy = kv_idx >= ctx_len

# teacher forcing mask logic
teacher_forcing = (
    # noisy queries attend to same-frame noisy keys
    (q_is_noisy & kv_is_noisy & (q_frame == kv_frame))
    # noisy queries attend to earlier clean keys
    | (q_is_noisy & ~kv_is_noisy & (q_frame > kv_frame))
    # clean queries attend causally to clean keys
    | (~q_is_noisy & ~kv_is_noisy & (q_frame >= kv_frame))
)

# sliding window constraint
sliding_window = kv_frame > (q_frame - L_s)

mask = teacher_forcing & sliding_window
\end{lstlisting}
\end{algorithm}

\end{document}